\documentclass{article}

\usepackage[nonatbib,preprint]{neurips_2020}
\usepackage{graphicx}
\usepackage{epstopdf}
\usepackage{mathptmx}      
\usepackage{subfigure}
\usepackage{amsmath}
\allowdisplaybreaks[4]
\usepackage{hyperref}

\usepackage{amsfonts}

\usepackage[ruled]{algorithm2e}

\usepackage{bm}

\usepackage{multirow}
\usepackage{enumerate}
\usepackage{amssymb}
\usepackage{appendix}
\usepackage{makecell}

\newtheorem{theo}{\textbf{Theorem}}
\newtheorem{def1}{\textbf{Definition}}

\newtheorem{lem}[theo]{\textbf{Lemma}}
\newtheorem{ass}{\textbf{Assumption}}
\newtheorem{cor}[theo]{\textbf{Corollary}}
\newtheorem{rem}{\textbf{Remark}}[section]
\newcommand*{\QEDA}{\hfill\ensuremath{\blacksquare}}%

\begin{document}

\title{Second-Order Convergence of Asynchronous Parallel Stochastic Gradient Descent: When Is the Linear Speedup Achieved?}

\author{Lifu Wang\\
  Beijing Jiaotong University\\
  \texttt{Lifu\_Wang@bjtu.edu.cn} \\
  \And
  Bo Shen\\
  Beijing Jiaotong University\\
  \texttt{bshen@bjtu.edu.cn} \\
  \And
  Ning Zhao\\
  Beijing Jiaotong University\\
  \texttt{n\_Zhao@bjtu.edu.cn}
}

\maketitle
\begin{abstract}
In machine learning, asynchronous parallel stochastic gradient descent (APSGD) is broadly used to speed up the training process through multi-workers. Meanwhile, the time delay of stale gradients in asynchronous algorithms is generally proportional to the total number of workers, which brings additional deviation from the accurate gradient due to using delayed gradients. This may have a negative influence on the convergence of the algorithm. One may ask: How many workers can we use at most to achieve a good convergence and the linear speedup?

In this paper, we consider the second-order convergence of asynchronous algorithms in non-convex optimization. We investigate the behaviors of APSGD with consistent read  near strictly saddle points and provide a theoretical guarantee that if the total number of workers is bounded by $\widetilde{O}(K^{1/3}M^{-1/3})$ ($K$ is the total steps and $M$ is the mini-batch size), APSGD will converge to good stationary points ($||\nabla f(x)||\leq \epsilon, \nabla^2 f(x)\succeq  -\sqrt{\epsilon}\bm{I}, \epsilon^2\leq O(\sqrt{\frac{1}{MK}}) $)  and the linear speedup is achieved. Our works give the first theoretical guarantee on the second-order convergence for asynchronous algorithms. The technique we provide can be generalized to analyze other types of asynchronous algorithms to understand the behaviors of asynchronous algorithms in distributed asynchronous parallel training.
\end{abstract}
\section{Introduction}
In large scale machine learning optimization problems, Stochastic Gradient Descent (SGD) has been widely used, whose convergence rate is $O(1/\sqrt{K})$ where $K$ is the number of steps. When there are multiple workers, it is feasible to speed up the convergence rate by parallel stochastic gradient methods. In \cite{JMLR:v13:dekel12a}, the authors studied the synchronous parallel stochastic gradient method, which uses multiple workers to compute the stochastic gradient on $M$ data in parallel and performs a synchronization before modifying parameters, and the convergence rate $O(1/\sqrt{MK})$ is achieved. However, since the synchronization process is very time-consuming, the synchronous parallel stochastic gradient method is relatively inefficient. It is found feasible to use the asynchronous method, which allows all workers to work independently and does not need a global synchronization. The asynchronous parallelism has been successfully applied in speeding up the training process in \cite{dean2012large,agarwal2011distributed,yun2014nomad,zhang2014deep}.

In asynchronous algorithms, parameters are updated by stale gradients. For a distributed system with fixed processing speed, the more workers are used, the larger the time delay of stale gradients is. Generally, the time delay bound $T$ is proportional to the total number of workers\cite{Lian2015asynchronous}. When $T$ is too large, the convergence will be significantly affected. Thus there must be an upper bound on the total workers for a distributed asynchronous training system. Therefore a question arises naturally:
\begin{center}
\fbox{
  {\shortstack[l]{What is the upper bound of the number of workers we can use to achieve the linear speed up?
}}
}
\end{center}
This problem was studied  in the convex case  \cite{agarwal2011distributed} and the non-convex case\cite{Lian2015asynchronous}. The results in  \cite{Lian2015asynchronous} is that let $K$ be the total steps and $M$ be the mini-batch size. Set the learning rate $\eta\sim \sqrt{1/MK}$. If the total number of workers is bounded by $O(\sqrt{K/M})$, the convergence rate of $\ ||\nabla f(x)||^2\ $ achieves $O(\sqrt{\frac{1}{MK}})$, thus the linear speedup is achieved.

However, all the previous results only provide a guarantee on $||\nabla f(x)||\to 0$. In the non-convex optimization, due to the existence of the saddle points, the asynchronous algorithm may converge to saddle points that are not local minima, and saddle points can be very bad. On the other hands, local minima can be good enough in many practical problems,
such as  multi-layer linear neural networks \cite{Kawaguchi2016Deep},
matrix completion, matrix sensing, robust PCA \cite{Rong2016Matrix},
Burer-Monteiro style low rank optimization \cite{Zhu2018Global} and over-parametrization neural network \cite{Du2018On}.
In these problems, under good conditions, all the saddle points of the loss are strictly saddle, and there is only one local minimum. Meanwhile, based on the work in \cite{Rong2015Escaping,lee2016gradient,jin2017how,fang2019sharp,jin2019stochastic,Staib2019Escapin}, it is well-known a perturbed SGD can escape strictly saddle points quickly and find a local minimum. The goal of this paper is to study that after adding a small noise, for asynchronous parallel SGD with consistent read, how large can the time delay be to keep the second-order convergence? Previous results\cite{Lian2015asynchronous} show that $T\leq O(K^{1/2}M^{-1/2})$ is enough for asynchronous SGD to converge to the first-order stationary points. In this paper, we show that when $T\leq \widetilde{O}(K^{1/3}M^{-1/3})$, asynchronous parallel SGD  does converge to the second-order stationary points ($||\nabla f(x)||\leq \epsilon, \nabla^2 f(x)\succeq  -\sqrt{\rho\epsilon}\bm{I}, \epsilon^2\leq O(\sqrt{\frac{1}{MK}}) $) and the linear speedup is achieved. Since second-order stationary points are almost local minima, this work illustrates that when and how we can obtain a model with a good performance by asynchronous training in non-convex optimization. A summary of the bounds for asynchronous algorithms is provided in Table \ref{tab1}.
\begin{table}[h]\label{tab1}
\caption{A summary of the bounds of number of workers for asynchronous algorithms.}
\begin{center}
\begin{tabular}{l l l l l}
\hline
Paper & Algorithm &Non-convex & \makecell[l]{Second Order\\ Convergence} & \makecell[l]{Number \\of Workers}\\
\hline
\makecell[l]{Agarwal and \\Duchi \cite{agarwal2011distributed}} & \makecell[l]{Asynchronous Parallel \\SGD with Consistent Read}
&No & N/A & $O(K^{1/4}M^{-3/4})$ \\
\hline
Liu et al. \cite{JMLR:v16:liu15a} &\makecell[l]{Asynchronous Parallel \\Stochastic Coordinate \\Descent(n is the \\coordinates dimension)}
& No & N/A & $O(n^{1/2})$ \\
\hline
Lian et al. \cite{Lian2015asynchronous} & \makecell[l]{ Asynchronous Parallel \\SGD with Consistent Read}
& Yes & No &  $O(K^{1/2}M^{-1/2})$ \\
\hline
This paper &\makecell[l]{Perturbed Asynchronous\\ Parallel SGD with\\ Consistent Read} & Yes & Yes & $\widetilde{O}(K^{1/3}M^{-1/3})$\\
\hline
\end{tabular}
\end{center}
\end{table}
\subsection{Our Contribution}
In this paper, we study the second-order convergence properties of the asynchronous stochastic gradient descent. We prove that the perturbed version of the asynchronous SGD algorithm will reach second-order stationary points in an almost dimension-free time, and the linear speedup is achieved.

Our main contributions are listed below:
\begin{itemize}
\item We design a novel approach to analyze the behaviors of asynchronous stochastic gradient descent near and far from strictly saddle points and study the influence of stale gradients in the updating. We establish an inequality to describe the special behaviors of asynchronous algorithms near saddle points and use  Lyapunov-Razumikhin methods for time-delay systems to study the instability. Our technique can also be generalized to analyze other types of asynchronous algorithms to understand the behaviors of asynchronous training, and our result reveals when and how asynchronous parallel stochastic gradient descent can speed up training without loss of the performance.
\item We study the influence of the time delay bound, in other words, the bound of number of the total workers on asynchronous training, and give the first theoretical guarantees for the asynchronous stochastic gradient descent algorithm to converge to second-order stationary points with a linear speedup (rate $O(\frac{1}{\sqrt{KM}})$)  when the total number of workers is upper bounded by $\widetilde{O}(K^{1/3}M^{-1/3})$. Thus we prove that in asynchronous parallel SGD, the workload for every worker can be reduced by the factor $\widetilde{O}(K^{1/3}M^{-1/3})$ compared to the synchronous parallel algorithms without causing any loss of accuracy and the ability to escape strictly saddle points.
\end{itemize}

\subsection{Related Works}
{\bf Asynchronous parallel SGD:} Asynchronous parallel SGD algorithm was firstly proposed in \cite{agarwal2011distributed}, and a lock-free version Hogwild was proposed in \cite{recht2011hogwild}. APSGD was used in Google to train deep learning networks effectively in \cite{dean2012large}.
The convergence was proved for convex cases in \cite{agarwal2011distributed,recht2011hogwild}, and non-convex cases were studied in \cite{Lian2015asynchronous},\cite{desa2015taming}. However, all these works are limited to first-order convergence.

{\bf First-order algorithms that escape saddle points:} The saddle escaping problem was firstly studied in \cite{Rong2015Escaping}. More detailed studies were given in \cite{jin2017how} for perturbed gradient descent and \cite{jin2019stochastic,fang2019sharp} for SGD. Stable manifold  in dynamical system was used in \cite{lee2016gradient} to show gradient descent will always finally reach a local minimum with probability almost one if we use random initialization. However, the work in \cite{Du2017Gradient} pointed out that if we don't add any noise, gradient descent may take exponential time to escape strict saddle points.

{\bf Stability of time-delay systems:} Saddle points escaping is closely related to the instability of the dynamical system. For the time-delay system, the stability has been studied in many work \cite{Zhou2018Improved,Gu1999Discretized,Han2005On,XU1994An,Kharitonov2003Lyapunov}.
Yet there are only a few articles about the instability of time-delay system, e.g.
\cite{Haddock1996Instability,SEDOVA20102324,Hale1965Sufficient,Raffoul2013INEQUALITIES}, which use Lyapunov-Krasovskii functional and Lyapunov-Razumikhin methods. In this paper, we prove a much stronger Razumikhin type instability theorem than previous results, which is available for asynchronous algorithms.

{\bf Notation} We use asymptotic notations $O(\cdot)$, $\widetilde{O}(\cdot)$, where $O(\cdot)$ is the general big O notation, and we use $\widetilde{O}(\cdot)$ to hide the logarithmic factors in $O(\cdot)$. $x\in \mathbb{R}^d$ denotes the parameters to be trained, $d$ is the dimension, and $x_*$ denotes the global optimal solution. $||\cdot||$ denotes the 2-norm of a matrix. We use $x\sim y$ to denote there is a constant $0<b\leq \widetilde{O}(1)$ such that $x=by$.
\section{Preliminaries on Asynchronous Parallel Stochastic Gradient Descent in Computer Network}\label{sm}
In this paper, we consider using asynchronous stochastic gradient descent to solve
\begin{equation}\label{ef}
\min_{x\in \mathbb{R}^d} f(x)=\frac{1}{n}\sum_if_i(x),
\end{equation}
where $f_i(x)$ is smooth and can be non-convex.

There are two types of asynchronous parallel implementations of stochastic gradient descent. One is the asynchronous parallel SGD with consistent read for multiple workers in the computer network \cite{agarwal2011distributed} and asynchronous parallel SGD with inconsistent read for the shared memory system\cite{recht2011hogwild}. In this paper, we only focus on the asynchronous parallel SGD with consistent read (AsySGD-con).

Consider a network with the star-shaped topology, and the center in the star-shaped network is the master machine. Node machines in the computer network only need to exchange information with the master machine. The process of perturbed asynchronous stochastic gradient descent in computer network is shown in algorithm \ref{a2}.
\begin{algorithm}\label{a2}
\caption{Perturbed Asynchronous Parallel Stochastic Gradient Descent in Computer Network}
{\bf Input:} Initial parameters $x_0$, learning rate $\eta$, perturbation radius $r$.

{\bf At the master machine:}\\
At time t, wait till receiving M stochastic gradients $g(x_{t-\tau_{t,i}},\theta_{t,i})$ from node machines.\\
$x_{t+1}=x_t-\eta (\sum_{i=1}^M g(x_{t-\tau_{t,i}},\theta_{t,i})+\sqrt{M}\zeta_t)$,\ \ \ $\zeta_t \sim N(\bm{0},(r^2/d)\bm{I})$. For  all $t, i, $ $0\leq \tau_{t,i}\leq T$;

{\bf At node machines:}\\
Pull parameter $x$ from the master.\\
Random select a sample $\theta_{t,i}$, compute stochastic gradient indexed by $ g(x,\theta_{t,i}))$ and push $g$ to the master machine.
\end{algorithm}

In this perturbed asynchronous SGD, an isotropic noise $\sqrt{M}\zeta_t, \zeta_t \sim N(\bm{0},(r^2/d)\bm{I})$ is added. As shown in \cite{Du2017Gradient}, this condition is necessary, otherwise gradient descent can take exponential time to escape strictly saddle points if we don't make any assumptions on the stochastic gradients as in \cite{fang2019sharp,lucchi2018escaping}.
\section{Second-order Convergence Guarantee of Asynchronous Parallel Stochastic Gradient Descent}
In this section we show when $T\leq \widetilde{O}(K^{1/3}M^{-1/3})$, Algorithm \ref{a2} will find reach a second-order stationary point. Our result is based on the following standard assumptions.
\begin{ass}\label{ass1}
Function $f(x)$ should be $L$ smooth and $\rho$-Hessian Lipschitz:
\begin{equation}
||\nabla f(x)-\nabla f(y)||\leq L ||x-y||\text{ , }||\nabla^2 f(x)-\nabla^2 f(y)||\leq \rho ||x-y|| \ \ \forall x,y.
\end{equation}
\end{ass}
\begin{ass}\label{ass3}
Stochastic gradient $g(x,\theta)$ should be $s^2$-norm-subGaussian:
\begin{equation}
\mathbb{E}g(x,\theta)=\nabla f(x), \ \ \ P(||g(x,\theta)-\nabla f(x)||\geq t)\leq 2exp(-t^2/(2s^2)),
\end{equation}
and $\sum_{i=1}^M g(x_{t-\tau_{t,i}},\theta_{t,i})- \nabla f(x_{t-\tau_{t,i}})$ is $Ms^2$-norm-sub-Gaussian.
\end{ass}
\begin{ass}\label{a4}(Batch Stochastic Gradient Assumption)
For any $\theta$, stochastic gradient function $g(\cdot,\theta)$ is $\ell$-Lipschitz.
\end{ass}
When we use minibatch-SGD with the random sampling, the gradient function $g(\cdot,\theta)$ is  $\ell$-Lipschitz for some $\ell$ if $f_i(x)$ in (\ref{ef}) are Lipschitz for all $i$. Then this assumption is true \cite{fang2019sharp,jin2019stochastic}.

In algorithm \ref{a2}, we denote $\sum_{m=1}^M\zeta_{j,m}=\sum_{i=1}^M (g(x_{t-\tau_{t,i}})- \nabla f(x_{t-\tau_{t,i}}))+\sqrt{M}\zeta_t$. Then $\sum_{m=1}^M\zeta_{j,m}$ is an $M\sigma^2$ norm-sub-Gaussian random vector, where $\sigma^2=s^2+r^2$. We set
\begin{equation}
\begin{aligned}
&\eta=\frac{\epsilon^2}{w\sigma^2L} \text{ with } w\leq \widetilde{O}(1) \text{ , } F=60c\sigma^2 \eta LT \leq T\epsilon^2 \text{ , } F_2=T_{max}\eta L\sigma^2\\
&T_{max}=T+\frac{u e^{(T+1)M \eta\sqrt{\rho\epsilon}/2}}{M\eta \sqrt{\rho \epsilon}/2}  \text{ , } 2\eta^2M^2L^2T^3\leq 1/5 \text{ , }  c=4
\end{aligned}
\end{equation}
where $0< u\leq \widetilde{O}(1)$.
Below we introduce a definition that is very useful for the analysis of asynchronous algorithms with delay $T\geq 1$.
\begin{def1}
Let $K$ be the total number of iterations.
We divide $K$ into $\lceil K/2T \rceil$ blocks $S_k=\{i|k2T\leq i<(k+1)2T\}$. \\
1) Blocks $S_k$ satisfying  $\sum_{i\in S_k}||\nabla f(x_i)||^2 \geq F$ are blocks of the first kind. \\
2) Let $F_k=\max_{i\in S_k}+1$. $S_k$ with  $\sum_{i\in S_k}||\nabla f(x_i)||^2 < F$, $\lambda_{min}(\nabla^2 f(x_{F_k}))\leq -\sqrt{\rho \epsilon}/2$ are blocks of the second kind.\\
3) Blocks are of the third kind, if $\sum_{i\in S_k}||\nabla f(x_i)||^2 < F$ and $\lambda_{min}(\nabla^2 f(x_{F_k}))> -\sqrt{\rho\epsilon}/2$.
\end{def1}

We are ready to present our main results on the bound of $T$ in Algorithm \ref{a2}.
\begin{theo}\label{m2}
Under the above assumptions, for a smooth function $f(x)$,
we run perturbed asynchronous parallel stochastic gradient algorithm \ref{a2} with $K$ iterations using parameter $r=s, \eta \sim \sqrt{\frac{1}{MLK}}$, $\epsilon^2\sim \sqrt{\frac{1}{MK}}$.
Suppose
\begin{equation}
T\leq \widetilde{O}(K^{1/2}M^{-1/2}),
\end{equation}
then with  high probability, asynchronous parallel stochastic gradient will reach points in the third kind of blocks at least once in $K$ iterations.
\end{theo}
\begin{theo}\label{m1}
Under the conditions of Theorem \ref{m2}, suppose
\begin{equation}
T\leq \widetilde{O}(K^{1/3}M^{-1/3}),
\end{equation}
then with  high probability, asynchronous parallel stochastic gradient will reach points $x$ satisfying $||\nabla f(x)||\leq \epsilon$ and $\lambda_{min}(\nabla^2 f(x))\geq -\sqrt{\rho\epsilon}$ at least once in $K$ iterations.
 \end{theo}
\begin{rem}
The bound of $T$ in this paper is $\widetilde{O}(K^{1/3}M^{-1/3})$, which is worse than that in the first order case \cite{Lian2015asynchronous} $T\leq  O(K^{1/2}M^{-1/2})$. However, we only need $T\leq \widetilde{O}(K^{1/2}M^{-1/2})$ to prove Theorem \ref{m2}, i.e. find a point $x\in S_k$ with
\begin{equation}\label{ss1}
\begin{aligned}
\sum_{i\in S_k}||\nabla f(x_i)||^2 < F \text{ , } \lambda_{min}(\nabla^2 f(x_{F_k}))> -\sqrt{\rho\epsilon}/2.
\end{aligned}
\end{equation}
For a block of the third kind, we have
\begin{equation}
||\nabla f(x_{i*})||^2=\min_{i\in\{k-T...k-1\}}||\nabla f(x_i)||^2\leq \frac{1}{T}\sum_{i=k-T}^{k-1}||\nabla f(x_i)||^2 < \frac{1}{T}F\leq \epsilon^2.
\end{equation}
And under the condition $T\leq \widetilde{O}(K^{1/3}M^{-1/3})$ in Theorem \ref{m1}, we can show
\begin{equation}
\lambda_{min}\nabla^2f(x_{i*})>-\sqrt{\rho\epsilon}/2-\rho ||x_{i*}-x_{k}||> -\sqrt{\rho\epsilon}/2- \sqrt{\rho\epsilon}/2\sqrt{\rho\epsilon}/L\geq -\sqrt{\rho\epsilon}.
\end{equation}
If we drop out this condition, (\ref{ss1}) will only provide a guarantee that algorithm \ref{a2} will find a point $x$ that either
\begin{equation}
\begin{aligned}
||\nabla f(x)||\leq \epsilon \text{ , } \lambda_{min}(\nabla^2f(x))\geq -\sqrt{(T+1)\rho\epsilon},
\end{aligned}
\end{equation}
or
\begin{equation}
\begin{aligned}
\ \ ||\nabla f(x)||\leq \sqrt{(T+1)}\epsilon \text{ , } \lambda_{min}(\nabla^2f(x))\geq -\sqrt{\rho\epsilon}.
\end{aligned}
\end{equation}
This is why Theorem \ref{m1} requires $T\leq \widetilde{O}(K^{1/3}M^{-1/3})$.
\end{rem}

Since the time delay parameter $T$ is generally proportional to the number of workers \cite{Lian2015asynchronous}, this theorem indicates that if the total number of workers is bounded by $\widetilde{O}(K^{1/3}M^{-1/3})$, the linear speedup is achieved to converge to a second-order stationary points $||\nabla f(x)||\leq \epsilon$ and $\lambda_{min}(\nabla^2 f(x))\geq -\sqrt{\rho\epsilon}$.
\section{Convergence Rate of APSGD to Reach the Third Kind of Blocks}
In this section, we show the main idea to prove Theorem \ref{m2}. Firstly, we consider the first-order convergence. The following theorem is a variant of Theorem 1 in \cite{Lian2015asynchronous}.
\begin{theo}\label{tl1}
Supposing $\eta^2(\frac{3 L}{4}-L^2M T^2\eta )-\frac{\eta}{2M}<0$, with probability at least $1-3e^{-\iota}$, we have
\begin{equation}
\begin{aligned}
f(x_{t_0+\tau+1})-f(x_{t_0}) &\leq \sum_{k=t_0}^{t_0+\tau} -\frac{3M\eta}{8}||\nabla f(x_k)||^2 + c\eta \sigma^2\iota +2\eta^2LM^2 c\sigma^2(\tau+1+\iota)\\
&+L^2T^2M\eta^3 \sum_{k=t_0-T}^{t_0-1} ||\sum_{m=1}^M\nabla f(x_{j-\tau_{j,m}})||^2.
\end{aligned}
\end{equation}
\end{theo}
\begin{rem}
When we set $t_0=0$, $\tau=K$, this theorem shows when $\eta^2(\frac{3 L}{4}-L^2M T^2\eta )-\frac{\eta}{2M}<0$, $f(x_K)-f(x_0)\leq \sum_{k=0}^{K} -\frac{3M\eta}{8}||\nabla f(x_k)||^2+ \text{some constants}$. Thus as in the synchronous case, we only need to show $\sum_k||\nabla f(x_k)||^2$ is large.
However, when $t_0>0$, the ``memory effect'' term  $L^2T^2M\eta^3 \sum_{k=t_0-T}^{t_0-1} ||\sum_{m=1}^M\nabla f(x_{j-\tau_{j,m}})||^2$ is important, and there is no guarantee that $f(x_k)$ will keep decreasing as $k$ increasing. This observation is crucial in the analysis on the behaviors near saddle points.
\end{rem}
Next we consider the behaviors near a strictly saddle point $x$ with  $||\nabla f(x)||\approx 0, \lambda_{min}(\nabla f(x))\leq -\gamma$ for some $\gamma>0$.
\begin{theo}\label{tl2} Supposing $\eta L MT\leq 1/3$, given a point $x_k$, let $\bm{H}=\nabla^2 f(x_k)$, and $e_1$ be the minimum eigendirection of $\bm{H}$,
$\gamma= -\lambda_{min}(\bm{H})\geq\sqrt{\rho \epsilon}/2$ and $\sum_{t=k-2T}^{k-1}||\nabla f(x_t)||^2\leq F$. We have, with probability at least $1/24$,
$$\sum_{t=k}^{k+T_{max}-1}||\nabla f(x(t))||^2\geq F_2=T_{max}\eta L\sigma^2.$$
\end{theo}
The main difference between asynchronous and synchronous case is that in the synchronous case, for any $x_k$ with $\lambda_{min}(\nabla^2 f(x_k))\leq -\sqrt{\rho\epsilon}/2$, $\sum_{t=k}^{k+T_{max}-1}||\nabla f(x(t))||^2$ will be large even if $||\nabla f(x_k)||$ is not small \cite{jin2017how}. However in the asynchronous case, $\sum_{t=k-2T}^{k-1}||\nabla f(x_t)||^2\leq F$ is necessary due to the stale gradients.

Then we have the following lemma, and Theorem \ref{m2} is a direct corollary of it.
\begin{lem}\label{lml}
For a large enough $\iota$, let
$$K=\max \{100\iota T\frac{f(x_0)-f(x_*)}{M\eta F},100\iota T_{max}\frac{f(x_0)-f(x_*)}{M\eta F_2}\}.$$
With probability at least $1-3e^{-\iota}$, we have:\\
$1)$ There are at most $\lceil K/8T \rceil$ blocks of the first kind .\\
$2)$ There are at most $\lceil K/8T \rceil$ blocks of the second kind.\\
so that at least $\lfloor K/4T \rfloor$ blocks are of the third kind.
\end{lem}
$1)$ is trivial because
\begin{equation}
\begin{aligned}
f(x_{\tau+1})-f(x_{0}) \leq \sum_{k=t_0}^{t_0+\tau} -\frac{3M\eta}{8}||\nabla f(x_k)||^2 + c\eta \sigma^2\iota +2\eta^2LM^2 c\sigma^2(\tau+1+\iota).
\end{aligned}
\end{equation}
For $2)$, let $z_i$ be the stopping time such that
\begin{equation}
\begin{aligned}
&z_1=\inf\{j| S_j \text{ is of the second kind}\},\\
&z_i=\inf\{j| T_{max}/2T\leq j-z_{i-1} \text{ and } S_j \text{ is of the second kind}\}.\\
\end{aligned}
\end{equation}
Let $N = max\{i|2T \cdot z_i+T_{max} \leq K\}$. Note that for $X_i=\sum_{k=F_{z_i}}^{F_{z_i}+T_{\max}-1} ||\nabla f(x_k)||^2$, $\mathbb{E} X_i\geq \frac{1}{24}F_2$ by Theorem \ref{tl2}. $\sum_i^N X_i$ is a submartingale. Using Azuma's inequality, $2)$ follows.

\section{Behaviors Near Strictly Saddle Points}\label{ss4}
Theorem \ref{tl2} is the key theorem in the second-order convergence. In this section, we study the behaviors of APSGD near a strictly saddle point and prove Theorem \ref{tl2}. Then main idea is to study the exponential instability of APSGD near a strictly saddle point and use an inequality to give a lower bound of $\sum_{t=k}^{k+T_{max}}||\nabla f(x_t)||^2$ to prove Theorem \ref{tl2}.
\subsection{Descent Inequality}
The behaviors of asynchronous gradient descent are quite different from the synchronous case in \cite{Rong2015Escaping,jin2017how}, which can be described by the following inequality:
\begin{lem}\label{loc2} Supposing $\eta^2(\frac{3 L}{4}-L^2M T^2\eta )-\frac{\eta}{2M}<0$ ,
\begin{equation}
\begin{aligned}
\sum_{k=t_0}^{t-1+t_0}(1+2L^2\eta^2M^2T^3) ||\nabla f(x_{k})||^2 &\geq\frac{||x_{t_0+t}-x_{t_0}||^2-3\eta^3||\sum_m \sum_{i=t_0}^{t_0+t-1}\zeta_{i,m} ||^2}{3\eta^2M^2t}\\
&-\sum_{k=t_0-2T}^{t_0-1}2L^2M^2\eta^2T^3||\nabla f(x_{k})||^2 \\
&-\sum_{k=t_0}^{t-1+t_0} 2L^2\eta^2||\sum_{j=k-\tau^{max}_{k}}^{k-1}\sum_{m=1}^M\zeta_{j,m}||^2,\\
\end{aligned}
\end{equation}
where $0\leq \tau^{max}\leq T$ is a random variable.
\end{lem}
If $\sum_{k=t_0-2T}^{t_0-1}||\nabla f(x_{k})||^2$ is very large, in the worst case, even if $\max_{t\leq T_{max}}||x_{k+t}-x_k||$ is large enough, $\sum_{k=t_0}^{t_0+2T}||\nabla f(x_{k})||^2$ can be still very small. This is
 due to there is no guarantee that the asynchronous gradient descent can decrease the function value, so it is possible that the algorithm will finally return back to a point near the
 saddle point. However, if $||\nabla f(x_{k})||^2$ keeps  small for a long enough time ($>2T$), we have $\sum_{k=t_0-2T}^{t_0-1}||\nabla f(x_{k})||^2\leq F$,
then   $\max_{t\leq T_{max}}||x_{k+t}-x_k||$ is large $\Rightarrow \sum_{k=t_0}^{t_0+T_{max}}||\nabla f(x_{k})||^2\geq F_2$ by Lemma \ref{loc2}.
Thus there is a direct corollary:
\begin{lem}\label{l3}
There is a parameter $S\sim \sqrt{L\eta MT_{max}}\eta \sqrt{M}\sqrt{T_{max}}\sigma$ such that, supposing $\eta L MT\leq 1/3$, if $\sum_{t=k-2T}^{k-1}||\nabla f(x_t)||^2\leq F $, we have
\begin{equation}
\begin{aligned}
P(\sum_{t=k}^{k+T_{max}-1}||\nabla f(x_t)||^2 \geq F_2,\  { or }\  \forall t\leq T_{max}, ||x_{k+t}-x_k||^2\leq S^2)\geq 1- 1/24.
\end{aligned}
\end{equation}
\end{lem}
Using this lemma, in order to prove Theorem \ref{tl2}, we can turn to show  $||x(t)-x(0)||>S$.
\subsection{Exponential Instability of Asynchronous Gradient Dynamics}\label{eia}
We will show $||x(t)-x(0)||>S$ by analyzing the exponential instability of asynchronous gradient dynamics near the strict saddle points. In fact we have:
\begin{theo}\label{st}
Supposing $x_k$ satisfying $\lambda_{min}(\nabla^2f(x_k))\leq-\sqrt{\rho\epsilon}/2$, we have
\begin{equation}
P(\max_{t\leq T_{max}}||x_{k+t}-x_k||\geq S)\geq 1/12.
\end{equation}
\end{theo}
To illustrate the main idea of the proof of Theorem \ref{st}, we provide a sketch.\\
\textit{Proof sketch}. As in \cite{jin2017how}, consider two sequences $\{x_1(t)\}$ and $\{x_2(t)\}$ as  two separate runs of APSGD starting from $x_k$ and for all $t\leq k$, $x_1(t)=x_2(t)$.
The Gaussian noise $\zeta_1(t)$ and $\zeta_2(t)$ in $\{x_1(t)\}$ and $\{x_2(t)\}$ satisfy $e_1^T\zeta_1=-e_1^T\zeta_2$, where $e_1$ is the eigenvector of $\lambda_{min}(\nabla^2f(x_k))$.
Other components at any direction perpendicular to $e_1$ of $\zeta_1$ and $\zeta_2$ are equal. Consider $x(t)=x_1(k+t)-x_2(k+t)$. We can prove that
\begin{equation}
\begin{aligned}
x(k)=&x(k-1)+\eta[\sum_{m=1}^M (\boldsymbol{H}+\Delta(k-\tau_{k,m}))x(k-\tau_{k,m}) +\sqrt{M}\hat{\zeta}_k +\sum_m \hat{\xi}_{k,m}],\\
\end{aligned}
\end{equation}
where $\bm{H}$ is a symmetric matrix with $\lambda_{max}(\bm{H})\geq\sqrt{\rho\epsilon}/2$, $||\Delta(k-\tau_{k,m})||\leq \rho S$, $\hat{\zeta}_k=2N(0,r^2/d)e_1$, $\hat{\xi}_{k,m}$ is $\ell^2||x(k)||^2$-norm-subGaussian by Assumption \ref{ass3} and \ref{a4}.

Then there is a polynomial function $f(t_0,t,y)$ such that $x(k)=\psi(k)+\phi(k)+\phi_{sg}(k)$
\begin{equation}
\begin{aligned}
&\psi(k)=\sqrt{M}\eta \sum_{i=0}^{k-1} f(i,k,\boldsymbol{H})\hat{\zeta}_i, \\
&\phi(k)=\eta \sum_m \sum_{i=0}^{k-1} f(i,k,\boldsymbol{H})\Delta(i-\tau_{i,m})x(i-\tau_{i,m}),\\
&\phi_{sg}(k)=\eta \sum_m\sum_{i=0}^{k-1} f(i,k,\boldsymbol{H})\hat{\xi}_{i,m},
\end{aligned}
\end{equation}
and $f(t_0,t,\boldsymbol{H})$ is the fundamental solution  of the following delayed linear equation
\begin{equation}\label{le}
\begin{aligned}
&x(k)=x(k-1)+\eta[\sum_{m=1}^m \boldsymbol{H}x(k-\tau_{k,m})],\\
&x(t_0)=\boldsymbol{I},\\
&x(n)=\boldsymbol{0} \text{ for all $n<t_0$}.
\end{aligned}
\end{equation}
Then we need to give a upper bound of $||\phi(k)+\phi_{sg}(k)||$ and a lower bound of $\psi(k)$. Following Lemma 6 in \cite{jin2019a}, we need to construct a matrix $\bm{Y}_\delta$
$$\boldsymbol{Y}_\delta=\begin{bmatrix}

       0 & (\phi+\phi_{sg})^T \\

       \phi+\phi_{sg} & 0

  \end{bmatrix}$$
and use Chernoff bound by estimating the bound of $\mathbb{E} tr\{ e^{\theta \boldsymbol{Y}_\delta }\}$. Using Chernoff bound arguments in \cite{jin2019a}, we can show that, with probability at least $1/6$,  $||\phi_{sg}(T_{max})+\phi(T_{max})||\leq \frac{\beta(T_{max})\sqrt{M}\eta r}{2\sqrt{d}}$ and $||\psi(T_{max})||\geq \frac{\beta(T_{max})2\sqrt{M}\eta r}{3\sqrt{d}}$. Thus $||x(T_{max})||\geq  \frac{\beta(T_{max})\sqrt{M}\eta r}{6\sqrt{d}}$, where $\beta^2(k)=\sum_{i=0}^k f^2(i,k,\sqrt{2\epsilon}).$

To estimate $\beta^2(k)$, we need to study the time-delayed equation (\ref{le}) using Razumikhin-Lyapunov method. Inspired by Mao-Razumikhin-Lyapunov stability theorem in stochastic differential equation with finite delay \cite{mao1996razumikhin,Mao1999Razumikhin}, we can prove a new instability theorem:
\begin{theo} \label{th8} For a discrete system,
$V(n,x)$ is a positive value Lyapunov function. Let $\Omega$ be the space of discrete function $x(\cdot)$ from $\{-T,...0,1,2,...\}$
to $\mathbb{R}$ and $x(\cdot)$ is a solution of the given discrete system equation. Suppose there exit $q, q_m$ satisfying the following two conditions
\begin{equation}\label{con}
\begin{aligned}
&(a)V(t+1,x(t+1))\geq q_m V(t,x(t)), q_m>0 \text{ (Bounded difference condition.) } \\
&(b)\text{If }\ V(t-\tau,x(t-\tau))\geq (1+q)^{-T} \frac{q_m}{1+q}V(t,x(t)) \forall\  0\leq \tau \leq T \\
&\text{then }\ V(t+1,x(t+1))\geq (1+q) V(t,x(t))\text{ (Razumikhin condition.) }
\end{aligned}
\end{equation}
Then for any $x(\cdot)\in \Omega$ satisfying  that for all $-T\leq t\leq 0 $, $V(t,x(t))\geq p V(0,x(0))$ with $0<p\leq 1$, we have $V(t,x(t))\geq (1+q)^t pV(0,x(0))$ for all $t>0$.
\end{theo}
Let $\boldsymbol{P}$ is the projection matrix to $e_1$. Apply this theorem to (\ref{le}) with $V(k,x)=||\boldsymbol{P}x(k)||$. In this case, $q_m=1$. Let $\gamma=\sqrt{\rho\epsilon}/2$. Condition (\ref{con}) $(b)$ of $q$ has the form
\begin{equation}
\begin{aligned}
1+M\eta \gamma(1+q)^{-T-1}\geq 1+q,\\
\end{aligned}
\end{equation}
which is equal to
\begin{equation}
\begin{aligned}
q (1+q)^{T+1}\leq M\eta\gamma.\\
\end{aligned}
\end{equation}
 It is easy to see that for all $T>0$, since $M\eta\gamma>0$, there is a $q>0$ satisfying Razumikhin condition (\ref{con}),
 so that the system is exponential unstable.
Let $T+1= \frac{f}{M\eta\gamma}$.
$$(1+q)^{T+1}=(1+q)^{\frac{f}{M\eta\gamma}}=(1+q)^{\frac{1}{q}\frac{q}{M\eta\gamma} f}\leq e^{f\frac{q}{M\eta\gamma}}\leq e^f$$
The last inequality is from $0<q\leq M\eta\gamma$. Thus $q\geq  e^{-f}M\eta\gamma$, $||\boldsymbol{P}x(n+1)||\geq (1+e^{-f}M\eta\gamma)|| \boldsymbol{P}x(n)||$ if $n>T$. Thus this theorem indicates that $f(k,t+1,\gamma) \geq (1+M\eta\gamma e^{-(T+1)M \eta\gamma} )f(k,t,\gamma)$ if $t-k \geq T$.
Then we have:
\begin{cor}\label{f1}
If $t-k\geq T$, $f(k,t+1)\geq (1+q) f(k,t)$, where $q=M\eta \gamma e^{-(T+1)M\eta\gamma}$.
\end{cor}
\begin{rem}
This theorem indicates that asynchronous algorithms will take $e^{(T+1)M\eta\gamma}$ times as long than the synchronization one to achieve $||x(t)-x(0)||>S$. In our case, $(T+1)M\eta\sqrt{\rho \epsilon}/2\leq O(1)$, $e^{(T+1)M\eta\gamma}\leq O(1)$ will not be too large.
\end{rem}
Combing these facts, we have $||x(T_{max})||>2S$ if $T_{max}=T+\frac{u e^{(T+1)M \eta\sqrt{\rho\epsilon}/2}}{M\eta \sqrt{\rho \epsilon}/2}$ and $u\leq \widetilde{O}(1)$. Note that
\begin{equation}
\begin{aligned}
\max(||x_1(k+T_{max})-x_1(k)||,||x_2(k+T_{max})-x_2(k)||)\geq \frac{1}{2}||x(T_{max})||.
\end{aligned}
\end{equation}
The theorem follows.
\section{Conclusion}\label{c1}
In this paper, we studied the theoretical properties of the popular asynchronous parallel stochastic gradient descent algorithm in non-convex optimization. We  gave the first theoretical guarantee that if the number of workers is bounded by $\widetilde{O}(K^{1/3}M^{-1/3})$, perturbed APSGD with consistent read will converge to a second-order stationary point($||\nabla f(x)||\leq \epsilon, \nabla^2 f(x)\succeq  -\sqrt{\rho\epsilon}\bm{I}$ ) and the linear speedup is achieved. Our results provide a  theoretical basis on when the asynchronous algorithms are effective in the non-convex case and take a step to understand the behaviors of asynchronous algorithms in distributed asynchronous parallel training.

\newpage
\section*{Broader Impact}
This is a theoretical paper. Broader Impact discussion is not applicable.
\bibliographystyle{unsrt}
\bibliography{References}

\newpage
{\bf \Large Supplementary Material}

\appendix

\section*{A Some Useful Lemmas}
In this section some lemmas and definitions  used in this paper are listed below.
\begin{def1}
We define the zero-mean nSG($\sigma_i$) sequence as the  sequence of random vectors $X_1,X_2....X_n\in \mathbb{R}^d$ with  filtrations $F_i=\sigma(X_1,X_2...X_i)$ such that
$$\mathbb{E}[X_i|F_{i-1}]=0, \mathbb{E} [e^{s||X_i||}|F_{i-1}]\leq e^{4s^2\sigma_i^2}, \sigma_i\in F_{i-1}$$.
\end{def1}
For a zero-mean nSG($\sigma_i$) sequence, we have some important lemmas from \cite{jin2019a}.

As in \cite{jin2019a}, for a zero-mean nSG $X_i$, let
$$\boldsymbol{Y}_i=\begin{bmatrix}

       0 & X^T_i \\

       X_i & 0

  \end{bmatrix}$$
and $c=4$, then we have:

\begin{lem}\label{pr}(Lemma 6 in \cite{jin2019a})
Supposing $\mathbb{E}tr\{e^{\sum_i\theta \bm{Y}_i}\}\leq e^{\sum\theta^2\sigma^2_i}(d+1)$, with probability at least $1-2(d+1)e^{-\iota}$:
$$||\sum_i X_i||\leq c\theta\sum_i^n\sigma^2_i+\frac{1}{\theta}\iota.$$
\end{lem}

\begin{lem}\label{ll1}
(SubGaussian Hoeffding inequality, Lemma 6 in \cite{jin2019a})
With probability at least $1-2(d+1)e^{-\iota}$:
$$||\sum_i^nX_i||\leq c\sqrt{\sum_i^n\sigma_i^2\cdot \iota}.$$
\end{lem}
The proof is based on Chernoff bound arguments and $\lambda(\bm{Y}_i)=0, ||X_i||$ or $-||X_i||$.

Using the same way, it is easy to prove the square sum theorem:
\begin{lem}\label{lsum}(Lemma 29 in \cite{jin2019stochastic})
For a zero-mean nSG($\sigma_i$) sequence $X_i$ with $\sigma_i=\sigma$, with probability at least $1-e^{-\iota}$:
$$\sum_i||X_i||^2\leq c\sigma^2(n+\iota).$$
\end{lem}

\begin{lem}\label{ex}
Let $\boldsymbol{X}$ be a $\sigma^2$-sub-Gaussian random vector, then
\begin{equation}
\mathbb{E} e^{\theta^2||\boldsymbol{X}||^2}\leq e^{65c\sigma^2\theta^2},
\end{equation}
if $\theta^2 \leq \frac{1}{16c\sigma^2}$.
\end{lem}
Note that $||\boldsymbol{X}||^2$ is a $16c \sigma^2$ sub-exponential random variable. This lemma can be proved by directly calculation:
$$\mathbb{E} e^{\theta^2||\boldsymbol{X}||^2}\leq e^{c\theta^2\sigma^2}(1+128c^2\theta^4\sigma^4)\leq e^{65c\theta^2\sigma^2}.$$

The following theorem is very useful in the proof of Chernoff bound:
\begin{lem} \label{ll8}
Let $\boldsymbol{Y}_i$ the random matrix  such that $\mathbb{E}\{\boldsymbol{Y}_i\}=0$ and $\mathbb{E}tr  \{ e^{\theta \boldsymbol{Y}_i}\} \leq e^{c\theta^2 \sigma_i^2}(d+1)$,
then we have
 $\mathbb{E}tr\{  e^{\theta \sum_i \boldsymbol{Y}_i}\} < e^{c\theta^2 (\sum_i\sigma_i)^2} (d+1)$.
\end{lem}

{\bf Proof:}

For any semi-positive definite matrix $\boldsymbol{A}_i$ and $\sum_i a_i=1 $, $a_i\geq 0$, we have \cite{chen} $$tr\prod_{i=1}^M \boldsymbol{A}_i^{a_i}\leq \sum_i a_i tr \boldsymbol{A}_i.\ \ \ $$
However it is impossible to use this inequality directly. When $\boldsymbol{Y}_i$ and $\boldsymbol{Y}_j$ are not commutative if $i\neq j$,
we have $e^{\sum_i^n \boldsymbol{Y}_i}\neq \prod_i^n e^{\boldsymbol{Y}_i} $ even $tr\{e^{\sum_i^n \boldsymbol{Y}_i}\}\neq tr\{\prod_i^n e^{\boldsymbol{Y}_i}  \}$.
In the case $n=2$, we have Golden-Thompson inequality \cite{Golden1965Lower} $tr\{e^{\boldsymbol{Y}_1+\boldsymbol{Y}_2}\}\leq tr\{e^{\boldsymbol{Y}_1}e^{\boldsymbol{Y}_2}  \}$.
However it is false when $n=3$, which is studied by Lieb in \cite{Lieb1973Convex}. Fortunately, for $n\geq3$, we have  Sutter-Berta-Tomamichel inequality \cite{Sutter2017Multivariate}:\\
Let $||\cdot ||$ be the trace norm, and $\boldsymbol{H}_k$ be Hermitian matrix. We have
\begin{equation}
\log ||exp(\sum_k^n \boldsymbol{H}_k)|| \leq \int \log ||\prod_k^n exp[(1+it)\boldsymbol{H}_k]||  d \beta(t),
\end{equation}
where $\beta$ is a probability measure.

For the right hand side, we have
$$||\prod_k^n exp[(1+it)\boldsymbol{H}_k]||\leq \sum_i \sigma_i(\prod_k^n exp[(1+it)\boldsymbol{H}_k) \leq  \sum_i \sigma_i(exp(\boldsymbol{H}_1))\sigma_i(exp(\boldsymbol{H}_2))...\sigma_i(exp(\boldsymbol{H}_n)),$$
where $\sigma_i$ is the ith singular value.

If all $H_k$ are semi-positive definite, $\lambda_i=\sigma_i$, using the elementary inequality that
$$\sum_i \lambda_i^{\alpha_1}(exp(\boldsymbol{H}_1))\lambda_i^{\alpha_2}(exp(\boldsymbol{H}_2))...\lambda_i^{\alpha_n}(exp(\boldsymbol{H}_n))\leq \sum_i (\sum_k\alpha_k \lambda_i(exp(\boldsymbol{H}_k))),$$
where $\sum_i \alpha_i=1$, we have
\begin{equation}
||\prod_k^n exp[(1+it)\boldsymbol{H}_k]||\leq  tr \{\sum_k \alpha_k exp(\boldsymbol{H}_k)\},
\end{equation}
so that
\begin{equation}
tr \{exp(\sum_k^n \alpha_k \boldsymbol{H}_k)\} \leq \sum_k \alpha_k tr \{ exp(\boldsymbol{H}_k)\}.
\end{equation}
\QEDA

In this supplementary material, the parameters we used are listed below:
\begin{equation}\label{ed}
\begin{aligned}
&\eta=\frac{\epsilon^2}{w\sigma^2L}\ \text{, }\ r=s \ \text{, }\ f=(T+1)M\eta\sqrt{\rho\epsilon}/2 \text{, }\  T_{max}=T+\frac{u e^{f}}{M\eta \sqrt{\rho \epsilon}/2} \ \text{, }\ c=4, \ \text{, } B\leq \tilde{O}(1),\\
&F=60c\sigma^2 \eta LT \leq T\epsilon^2\text{ , } F_2=T_{max}\eta L\sigma^2\ \text{ , } S=B\sqrt{L\eta MT_{max}}\eta \sqrt{M}\sqrt{T_{max}}\sigma \ \text{ , } w\leq \tilde{O}(1), \\
&b=\log(2(d+1)) +\log 2\ \text{, }\ \frac{2\sqrt{48c}+2b }{C}=\frac{1}{2}\ \text{, }\ p=\frac{1}{1+C} \ \text{, }c_2=\log96 +\log (d+1).
\end{aligned}
\end{equation}

\begin{lem}\label{ie}
We have $\eta\sim \epsilon^2, T_{max}\sim \frac{\epsilon^{-2.5}}{M}, F\sim T\epsilon^2, F_2\sim T_{max}\epsilon^{2}, S\sim M\epsilon^{0.5},w \leq \widetilde{O}(1), u \leq \widetilde{O}(1) $ such that  the following conditions are satisfied.
\begin{enumerate}[(a)]
\item $\eta \leq \frac{1}{3ML(T+1)}$ and $2\eta^2M^2L^2T^3\leq 1/5$
\item $\sqrt{3\times65}(M T_{max} \eta\rho S+ \sqrt{T_{max}M} \eta\ell) \leq p$
\item $ \frac{S^2-3\eta^2M\sigma^2 T_{max}c^2c_2}{3\eta^2M^2T_{max}}-2L^2\eta^2T^3M^2 F- c2T_{max}2L^2M\eta^2T\sigma^2\geq 2T F_2$
\item Let $q=M\eta\sqrt{\rho\epsilon}/2e^{-(T+1)M\eta\sqrt{\rho\epsilon}/2}$. We have $\frac{2^{u }\sqrt{M}\eta r}{6\sqrt{3}\sqrt{2q d}}\geq 2S$
\item $e^{-T_{max}+\log T+\log T_{max}}\leq 1/48$
\end{enumerate}
\end{lem}

{\bf Proof of Lemma \ref{ie}:}

Firstly,
\begin{equation}
\eta=\frac{\epsilon^2}{w\sigma^2L}
\end{equation}
Thus if $\epsilon^2\leq O(\frac{1}{MT^{2/3}})$, $2\eta^2M^2L^2T^3\leq 1/5$ $\eta L M (T+1) \leq O(1)$. $(a)$ follows.

For $(b)$, we have
\begin{equation}
T^2_{max} M^2\eta^2\rho^2 S^2\sim T^2_{max} M^2\eta^2\rho^2 L\eta MT_{max}\eta^2 M T^2_{max}\sigma^2= T^4_{max} \eta^5 M^4\rho^2 L \sigma^2\leq \tilde{O}(\frac{\eta L\sigma^2}{\epsilon^2})\leq O(\frac{1}{w})
\end{equation}
Thus if $w\leq \tilde{O}(1)$ is large enough, $T_{max} M\eta\rho S\leq O(1)$.

\begin{equation}
T_{max}M \eta^2\ell^2\leq O(\frac{\epsilon^2  \ell^2}{w\sigma^2 L\sqrt{\rho\epsilon}})
\end{equation}
$\sqrt{T_{max}M} \eta\ell\leq O(\epsilon^{3/4}/\sqrt{w})$. Thus if $w\leq \tilde{O}(1)$, $(b)$ follows.

As for $(c)$, note that $S^2=B^2L\eta MT_{max}\eta^2 M T^2_{max}\sigma^2$
\begin{equation}
\begin{aligned}
 &\frac{S^2-3\eta^2M\sigma^2 T_{max}c^2c_2}{3\eta^2M^2T_{max}}-2L^2\eta^2T^3M^2 F- c2T_{max}2L^2M\eta^2T\sigma^2\\
 &=\frac{B^2}{3}L\eta T_{max}^2\sigma^2 -\frac{1}{M}c^2c_2\sigma^2 -(2L^2\eta^2T^2M^2) T60c\sigma^2 \eta LT- (\eta TM L) 4c T_{max}L\eta \sigma^2\\
 &\geq \frac{B^2}{3}L\eta T_{max}^2\sigma^2 -\frac{1}{M}c^2c_2\sigma^2- 40c\sigma^2 \eta LT^2-  4/3c T_{max}L\eta \sigma^2
\end{aligned}
\end{equation}
Since $T_{max}>T$, and $L\eta T_{max}^2\sigma^2> \sigma^2 \eta LT^2$, $L\eta T_{max}^2\sigma^2> T_{max}L\eta \sigma^2$,
 there exist $B\leq O(1)$ such that $$ \frac{B^2}{3}L\eta T_{max}^2\sigma^2 -\frac{1}{M}c^2c_2\sigma^2- 40c\sigma^2 \eta LT^2-  4/3c T_{max}L\eta \sigma^2 \geq 2T T_{max}\eta L\sigma^2=2TF_2$$
Thus $(c)$ follows. \\
$2^{u }=\frac{6\sqrt{3}\sqrt{2q d} 2S}{\sqrt{M}\eta r}$, thus $u\leq \tilde{O}(1)$ and $(d)$ follows. When $u\leq \widetilde{O}(1)$ is large enough, $(e)$ follows.
\QEDA

\section*{B Proof of Theorem \ref{m1}}
In this section we prove Theorem \ref{m1}. Lemma \ref{lml} shows that with high probability at least $K/2$  iterations are of the third kind, such that there is a $k$ satisfying $\sum_{i=k-2T}^{k-1}||\nabla f(x_i)||^2 < F$ and $\lambda_{min}(\nabla^2 f(x_{k}))> -\sqrt{\rho \epsilon}/2$, then $$||\nabla f(x_{i*})||^2=\min_{i\in\{k-T...k-1\}}||\nabla f(x_i)||^2\leq \frac{1}{T}\sum_{i=k-T}^{k-1}||\nabla f(x_i)||^2 < \frac{1}{T}F\leq \epsilon^2.$$

Using $2\eta^2M^2L^2T^3\leq 1/5$, for every block of the third kind, with probability at least $1/10$,
\begin{equation}\label{e10}
\begin{aligned}
||x_{i*}-x_k||^2= &\eta^2||\sum_{i=i*}^{k-1}\sum_{m=1}^M\nabla f(x_{k-\tau_{k,m}})+\sum_m \sum_{i=i*}^{k-1}\zeta_{i,m}||^2\\
&\leq 2\eta^2MT\sum_{i=i*}^{k-1}\sum_m||\nabla f(x_{i-\tau_{i,m}})||^2 +2\eta^2||\sum_{i=i*}^{k-1}\sum_m \zeta_{i,m}||^2\\
&\leq 2\eta^2M^2 T^2F+20c\eta^2MT\sigma^2\\
&\leq \epsilon^2/4L^2
\end{aligned}
\end{equation}
And
\begin{equation}\label{ss2}
\lambda_{min}\nabla^2f(x_{i*})>-\sqrt{\rho\epsilon}/2-\rho ||x_{i*}-x_{k}||> -\sqrt{\rho\epsilon}/2- \sqrt{\rho\epsilon}/2\sqrt{\rho\epsilon}/L\geq -\sqrt{\rho\epsilon}.
\end{equation}
There are at least $\lfloor K/4T \rfloor$ blocks which are of the third kind. Using Hoeffding's lemma, with a high probability, we can find such $x_{i*}$.

Note that
$K=\max \{100\iota \frac{f(x_0)-f(x_*)}{M\eta \frac{F}{T}},100\iota \frac{f(x_0)-f(x_*)}{M\eta \frac{F_2}{T_{max}}}\}\sim \frac{1}{M\epsilon^4}$.
From
$2\eta^2M^2L^2T^3\leq 1/5$, we have $T\leq \widetilde{O}(K^{1/3}M^{-1/3})$.
Our theorem follows.
\QEDA

\section*{C Proof of Lemma \ref{lml}}

Using Theorem \ref{tl1}, if there are more than $\lceil K/8T \rceil$  blocks of the first kind, with probability$1-3e^{-\iota}$
\begin{align}
f(x_{K+1})-f(x_{0})&\leq \sum_{k=0}^{K} -\frac{3M\eta}{8}||\nabla f(x_k)||^2+ c\eta \sigma^2\iota  +2\eta^2LM c\sigma^2(K+1+\iota)\\ \notag
&\leq \sum_i\sum_{k\in S_i} -\frac{3M\eta}{8}||\nabla f(x_k)||^2 + c\eta \sigma^2\iota  +2\eta^2LM c\sigma^2(K+1+\iota)\\ \notag
&\leq -\frac{K}{8T} \frac{3M\eta}{8} F + c\eta \sigma^2\iota  +2\eta^2LM c\sigma^2(K+1+\iota)\\ \notag
&= -\frac{K}{8T} \frac{3M\eta}{8} (F-\frac{128\eta LT}{3}c\sigma^2) + c\eta \sigma^2\iota  +2\eta^2LM c\sigma^2(K+1+\iota)\\ \notag
&\leq -\frac{K}{8T} \frac{3M\eta}{8} (F-50c\sigma^2 \eta LT)+ c\eta \sigma^2\iota  +2\eta^2LM c\sigma^2(K+1+\iota)\\ \notag
&\leq -\frac{K}{8T} \frac{3M\eta}{8} \frac{1}{6}F+ c\eta \sigma^2\iota  +2\eta^2LM c\sigma^2(K+1+\iota). \notag
\end{align}
Since $K\geq100\iota T\frac{f(x_0)-f(x_*)}{M\eta F }$, and $\iota$ is large enough, it can not be achieved.

As for $2)$, let $z_i$ be the stopping time such that
\begin{equation}
\begin{aligned}
&z_1=\inf\{j| S_j \text{ is of second kind}\}\\
&z_i=\inf\{j| T_{max}/2T\leq j-z_{i-1} \text{ and } S_j \text{ is of second kind}\}.\\
\end{aligned}
\end{equation}
Let $N = max\{i|2T \cdot z_i+T_{max} \leq K\}$. We have
\begin{equation}\label{me}
\begin{aligned}
f(x_{K+1})-f(x_0)&\leq\sum_{k=0}^{K} -\frac{3M\eta}{8}||\nabla f(x_k)||^2 + c\eta \sigma^2\iota \\
&+\frac{3\eta^2L}{2}Mc\sigma^2(K+1+\iota)+ L^2 T M\eta^3 Mc\sigma^2(K+1+T+\iota)\\
&\leq c\eta \sigma^2\iota  +\frac{3\eta^2L}{2}Mc\sigma^2(K+1+\iota)+ L^2 T M\eta^3 Mc\sigma^2(K+1+T+\iota)\\
&+ \sum_i^N \sum_{k=F_{z_i}}^{F_{z_i}+T_{\max}-1}  -\frac{3M\eta}{8}||\nabla f(x_k)||^2.\\
\end{aligned}
\end{equation}
Let $$X_i= \sum_{k=F_{z_i}}^{F_{z_i}+T_{\max}-1} ||\nabla f(x_k)||^2$$
$\sum_i X_i$ is a submartingale and the last term of Eq.(\ref{me}) is $-\sum_i X_i$. Using Theorem\ref{tl2}, $P(X_i\geq F_2)\geq 1/24$. Let $Y_i$ be a random variable, such that $Y_i=X_i$ if $X_i\leq F_2$ else $Y_i=F_2$. Then we have a bounded sub-martingale $0\leq Y_i\leq X_i$. Using Azuma's inequality, we have
\begin{equation}
P(\sum_i^N X_i \geq \mathbb{E}\{\sum_i Y_i\} -\lambda)\geq P(\sum_i^N Y_i \geq \mathbb{E}\{\sum_i Y_i\} -\lambda)\geq 1-2e^{-\frac{\lambda^2}{2F_2^2N}}
\end{equation}
And it easy to see $\mathbb{E}\{\sum_i^N Y_i\}\geq  \frac{1}{24}NF_2$. We have
\begin{equation}
P(\sum_i^N Y_i\geq \frac{1}{24}NF_2-\sqrt{2N}F_2\sqrt{\iota})\geq 1-e^{-\iota}
\end{equation}
If there are more then $K/8T$ blocks of the second kind, we have $N\geq K/4T_{max}$.
$$\frac{1}{24}N-\sqrt{2N}\sqrt{\iota}\geq \frac{1}{48}N$$
With probability at least $1-3e^{-\iota}$
\begin{equation}
\begin{aligned}
f(x_{K+1})-f(x_0)&\leq c\eta \sigma^2\iota  +2\eta^2LMc\sigma^2(K+1+\iota)\\
& -\frac{3M\eta}{8}\frac{N}{48}F_2\\
\end{aligned}
\end{equation}
If $N\geq K/4T_{max}$, and $K\geq 100\iota T_{max}\frac{f(x_0)-f(x_*)}{M\eta F_2}$, it can not be achieved.
\QEDA
\section*{D Proof of theorem \ref{tl1}}
Firstly, we need a lemma:
\begin{lem}\label{lnm} Under the condition of theorem \ref{tl1}, we have
\begin{align}
f(x_{k+1})-f(x_k)&\leq  -\frac{M\eta}{2}||\nabla f(x_k)||^2+ (\frac{3\eta^2 L}{4}-\frac{\eta}{2M}) ||\sum_{m=1}^M\nabla f(x_{k-\tau_{k,m}})||^2\\ \notag
&+ L^2TM\eta \sum_{j=k-T}^{k-1} \eta^2||\sum_{m=1}^M\nabla f(x_{j-\tau_{j,m}})||^2
- \eta \left<\nabla f(x_k),\sum_{m=1}^M\zeta_{k,m}\right>\\ \notag &+\frac{3\eta^2L}{2}||\sum_{m=1}^M \zeta_{k,m}||^2+ L^2  M\eta^3 || \sum_{j=k-\tau^{max}_{k}}^{k-1}\sum_{m=1}^M \zeta_{j,m}||^2\\ \notag
\end{align}
\end{lem}
{\bf Proof of Lemma \ref{lnm}:}
This lemma is, in fact, a transformation of (30) in \cite{Lian2015asynchronous}.
\begin{align}
f(x_{k+1})-f(x_k)\leq &\left< \nabla f(x_k),x_{k+1}-x_k\right> +\frac{L}{2}||x_{k+1}-x_k||^2\\ \notag
=&-\left<\nabla f(x_k),\eta \sum_{m=1}^M \nabla f(x_{k-\tau_{k,m}})+\eta \sum_{m=1}^M\zeta_{k,m}\right>\\ \notag
&+\frac{\eta^2L}{2}||\sum_{m=1}^M\nabla f(x_{k-\tau_{k,m}})+\sum_{m=1}^M\zeta_{k,m}||^2\\  \notag
=&-\left<\nabla f(x_k),\eta \sum_{m=1}^M \nabla f(x_{k-\tau_{k,m}})\right> +\left<\nabla f(x_k),\eta\sum_{m=1}^M\zeta_{k,m}\right> \\ \notag &+\frac{\eta^2L}{2}||\sum_{m=1}^M\nabla f(x_{k-\tau_{k,m}})+\sum_{m=1}^M\zeta_{k,m}||^2\\ \notag
\overset{(1)}{=}& -\frac{M\eta}{2}(||\nabla f(x_k)||^2+ ||\frac{1}{M}\sum_{m=1}^M\nabla f(x_{k-\tau_{k,m}})||^2\\ \notag
&- ||\nabla f(x_k)-\frac{1}{M}\sum_{m=1}^M \nabla f(x_{k-\tau_{k,m}})||^2)- \eta \left<\nabla f(x_k),\sum_{m=1}^M\zeta_{k,m}\right>\\ \notag &+\frac{\eta^2L}{2}||\sum_{m=1}^M\nabla f(x_{k-\tau_{k,m}})+\sum_{m=1}^M \zeta_{k,m}||^2\\ \notag
=& -\frac{M\eta}{2}(||\nabla f(x_k)||^2+ ||\frac{1}{M}\sum_{m=1}^M\nabla f(x_{k-\tau_{k,m}})||^2- \underbrace{||\nabla f(x_k)-\frac{1}{M}\sum_{m=1}^M \nabla f(x_{k-\tau_{k,m}})||^2)}_{T_1}\\ \notag
&- \eta \left<\nabla f(x_k),\sum_{m=1}^M\zeta_{k,m}\right> +\frac{\eta^2L}{2}(\frac{3}{2}||\sum_{m=1}^M\nabla f(x_{k-\tau_{k,m}})||^2+3||\sum_{m=1}^M \zeta_{k,m}||^2)\\ \notag
\overset{(2)}{\leq}& -\frac{M\eta}{2}[||\nabla f(x_k)||^2+ ||\frac{1}{M}\sum_{m=1}^M\nabla f(x_{k-\tau_{k,m}})||^2\\ \notag
&-2L^2(||\sum_{j=k-\tau^{max}_{k}}^{k-1}\eta\sum_{m=1}^M \zeta_{j,m}||^2+ T\sum_{j=k-\tau^{max}_{k}}^{k-1} \eta^2||\sum_{m=1}^M\nabla f(x_{j-\tau_{j,m}})||^2)]\\ \notag
&(\tau^{max}_{k}=\arg \max_{m\in\{1,...M\}}||x_k-x_{\tau_{k,m}}||)\\ \notag
&- \eta \left<\nabla f(x_k),\sum_{m=1}^M\zeta_{k,m}\right> +\frac{\eta^2L}{2}(\frac{3}{2}||\sum_{m=1}^M\nabla f(x_{k-\tau_{k,m}})||^2+3||\sum_{m=1}^M \zeta_{k,m}||^2)\\ \notag
\leq& -\frac{M\eta}{2}[||\nabla f(x_k)||^2+ ||\frac{1}{M}\sum_{m=1}^M\nabla f(x_{k-\tau_{k,m}})||^2\\ \notag
&-2L^2  (\eta^2 ||\sum_{j=k-\tau^{max}_{k}}^{k-1}\sum_{m=1}^M \zeta_{j,m}||^2+ T\sum_{j=k-\tau^{max}_{k}}^{k-1} \eta^2||\sum_{m=1}^M\nabla f(x_{j-\tau_{j,m}})||^2)]\\ \notag
&- \eta \left<\nabla f(x_k),\sum_{m=1}^M\zeta_{k,m}\right> +\frac{\eta^2L}{2}(\frac{3}{2}||\sum_{m=1}^M\nabla f(x_{k-\tau_{k,m}})||^2+3||\sum_{m=1}^M \zeta_{k,m}||^2)\\ \notag
\leq& -\frac{M\eta}{2}||\nabla f(x_k)||^2+ (\frac{3\eta^2 L}{4}-\frac{\eta}{2M}) ||\sum_{m=1}^M\nabla f(x_{k-\tau_{k,m}})||^2\\ \notag
&+ L^2TM\eta \sum_{j=k-T}^{k-1} \eta^2||\sum_{m=1}^M\nabla f(x_{j-\tau_{j,m}})||^2
- \eta \left<\nabla f(x_k),\sum_{m=1}^M\zeta_{k,m}\right>\\ \notag &+\frac{3\eta^2L}{2}||\sum_{m=1}^M \zeta_{k,m}||^2+ L^2  M\eta^3 || \sum_{j=k-\tau^{max}_{k}}^{k-1}\sum_{m=1}^M \zeta_{j,m}||^2\\ \notag
\end{align}

In (1) we use the fact $\left<a,b\right>=\frac{1}{2}(||a||^2+||b||^2-||a-b||^2)$, and (2) is from the estimation of $T_1$ in  \cite{Lian2015asynchronous}.

From this lemma, we can observe that, different from the general SGD, since the stale gradients are used, there is no guarantee that the function value will decrease in every step. However, it can be proved that the overall trend of the function value is still decreasing.

{\bf Proof of theorem \ref{tl1}:}

\begin{align}
f(x_{t_0+\tau+1})&-f(x_{t_0})=\sum_{k=t_0}^{t_0+\tau} f(x_{k+1})-f(x_{k})\\ \notag
\leq &\sum_{k=t_0}^{t_0+\tau} -\frac{M\eta}{2}||\nabla f(x_k)||^2+ (\frac{3\eta^2 L}{4}-\frac{\eta}{2M}) ||\sum_{m=1}^M\nabla f(x_{k-\tau_{k,m}})||^2\\ \notag
&+ L^2TM\eta \sum_{j=k-T}^{k-1} \eta^2||\sum_{m=1}^M\nabla f(x_{j-\tau_{j,m}})||^2 \\ \notag
&- \eta \left<\nabla f(x_k),\sum_{m=1}^M\zeta_{k,m}\right> +\frac{3\eta^2L}{2}||\sum_{m=1}^M \zeta_{k,m}||^2+ L^2  M\eta^3 ||\sum_{j=k-\tau^{max}_{k}}^{k-1}\sum_{m=1}^M \zeta_{j,m}||^2 \\ \notag
= &\sum_{k=t_0}^{t_0+\tau} -\frac{M\eta}{2}||\nabla f(x_k)||^2+ \sum_{k=t_0}^{t_0+\tau} (\frac{3\eta^2 L}{4}-\frac{\eta}{2M}) ||\sum_{m=1}^M\nabla f(x_{k-\tau_{k,m}})||^2\\ \notag
&+ L^2TM\eta \sum_{j=k-T}^{k-1} \eta^2||\sum_{m=1}^M\nabla f(x_{j-\tau_{j,m}})||^2\\ \notag
&-\sum_{k=t_0}^{t_0+\tau}  \eta \left<\nabla f(x_k),\sum_{m=1}^M\zeta_{k,m}\right> +\frac{3\eta^2L}{2}||\sum_{m=1}^M \zeta_{k,m}||^2+ L^2  M\eta^3 ||\sum_{j=k-\tau^{max}_{k}}^{k-1}\sum_{m=1}^M \zeta_{j,m}||^2\\ \notag
\leq &\sum_{k=t_0}^{t_0+\tau} -\frac{M\eta}{2}||\nabla f(x_k)||^2\\ \notag
&+ \sum_{k=t_0}^{t_0+\tau} (\eta^2(\frac{3 L}{4}+L^2M T^2\eta )-\frac{\eta}{2M}) ||\sum_{m=1}^M\nabla f(x_{k-\tau_{k,m}})||^2\\ \notag
&+ L^2TM\eta^3 \sum_{k=t_0-T}^{t_0-1} T ||\sum_{m=1}^M\nabla f(x_{j-\tau_{j,m}})||^2\\ \notag
& \underbrace{-\sum_{k=t_0}^{t_0+\tau} \eta \left< \nabla f(x_k),\sum_{m=1}^M\zeta_{k,m}\right>
+\sum_{k=t_0}^{t_0+\tau}\frac{3\eta^2L}{2}||\sum_{m=1}^M \zeta_{k,m}||^2+\sum_{k=t_0-T}^{t_0+\tau} L^2  M\eta^3 ||\sum_{j=k-\tau^{max}_{k}}^{k-1}\sum_{m=1}^M \zeta_{j,m}||^2}_{T_2}\\ \notag
\end{align}
\QEDA

In order to estimate $T_2$, we can use lemmas in \cite{jin2019stochastic}.

Let $\zeta_k=\frac{1}{M}\sum_{m=1}^M\zeta_{k,m}$. With probability $1-e^{-\iota}$, we have
\begin{equation}
-\sum_{k=t_0}^{t_0+\tau} \eta \left< M\nabla f(x_k),\zeta_{k}\right> \leq \frac{\eta M}{8}\sum_{k=t_0}^{t_0+\tau} ||\nabla f(x_k)||^2+ c\eta \sigma^2\iota
\end{equation}
This is from Lemma 30 in \cite{jin2019stochastic}.

With probability $1-e^{-\iota}$,
\begin{equation}
\begin{aligned}
\sum_{k=t_0}^{t_0+\tau}\frac{3\eta^2L}{2}||\sum_{m=1}^M \zeta_{k,m}||^2 \leq \frac{3\eta^2L}{2}Mc\sigma^2(\tau+1+\iota)
\end{aligned}
\end{equation}
And with probability at least $1-e^{-\iota/TML\eta +\log(\tau+T)+\log T}\geq 1-e^{-\iota} $(when $\tau$ is large enough),
\begin{equation}
\begin{aligned}
\sum_{k=t_0-T}^{t_0+\tau} L^2  M\eta^3 &||\sum_{j=k-\tau^{max}_{k}}^{k-1}\sum_{m=1}^M \zeta_{j,m}||^2\\
&\leq L^2 T M\eta^3 Mc\sigma^2(\tau/TM\eta+1+T+\iota)\leq \frac{\eta^2L}{2}Mc\sigma^2(\tau+1+\iota)
\end{aligned}
\end{equation}
We have, with  probability $1-3e^{-\iota}$,
$$T_1\leq \frac{\eta M}{8}\sum_{k=t_0}^{t_0+\tau} ||\nabla f(x_k)||^2+ c\eta \sigma^2\iota +2\eta^2LMc\sigma^2(\tau+1+\iota)$$

$\eta^2(\frac{3 L}{4}-L^2M T^2\eta )-\frac{\eta}{2M}<0$. With probability at least $1-3e^{-\iota}$,
\begin{equation}\label{e1}
\begin{aligned}
f(x_{t_0+\tau+1})-f(x_{t_0}) &\leq \sum_{k=t_0}^{t_0+\tau} -\frac{3M\eta}{8}||\nabla f(x_k)||^2 + c\eta \sigma^2\iota \\
&+2\eta^2LMc\sigma^2(\tau+1+\iota)\\
&+ L^2TM\eta^3 \sum_{k=t_0-T}^{t_0-1} T ||\sum_{m=1}^M\nabla f(x_{j-\tau_{j,m}})||^2
\end{aligned}
\end{equation}
The theorem follows.
\QEDA
\section*{E Proof of Lemma \ref{loc2}}
\begin{align}
|| x_{t_0+t}-x_{t_0} ||^2-&3\eta^2||\sum_m \sum_{i=t_0}^{t-1}\zeta_{i,m}||^2\\
=&\eta^2||\sum_{k=t_0}^{t-1+t_0}\sum_{m=1}^M\nabla f(x_{k-\tau_{k,m}})+\sum_m \sum_{i=t_0}^{t-1}\zeta_{i,m}||^2 -3\eta^2||\sum_m \sum_{i=t_0}^{t-1}\zeta_{i,m}||^2\\ \notag
\leq &3\eta^2\sum_{k=t_0}^{t-1+t_0} t[||M\nabla f(x_k)||^2+||M\nabla f(x_k)- \sum_{m=1}^M \nabla f(x_{k-\tau_{k,m}})||^2]\\ \notag
\overset{(a)}{\leq} &3\eta^2 \sum_{k=t_0}^{t-1+t_0} M^2t||\nabla f(x_{k})||^2 \\ \notag
&+3\eta^2t\sum_{k=t_0}^{t-1+t_0}M^22L^2\eta^2
[||\sum_{j=k-\tau^{max}_{k}}^{k-1}\sum_{m=1}^M\zeta_{j,m}||^2+T||\sum_{m=1}^M\nabla f(x_{j-\tau_{j,m}})||^2]\\ \notag
\leq &3\eta^2 \sum_{k=t_0}^{t-1+t_0} M^2t||\nabla f(x_{k})||^2 \\ \notag
&+3\eta^2t\sum_{k=t_0}^{t-1+t_0}\sum_{j=k-T}^{k-1} M^22L^2\eta^2T
||\sum_{m=1}^M\nabla f(x_{j-\tau_{j,m}})||^2\\ \notag
 &+3\eta^2t\sum_{k=t_0}^{t-1+t_0}\sum_{j=k-T}^{k-1} M^22L^2\eta^2||\sum_{j=k-\tau^{max}_{k}}^{k-1}\sum_{m=1}^M \zeta_{j,m}||^2\\ \notag
 \leq &3\eta^2 \sum_{k=t_0}^{t-1+t_0} M^2t||\nabla f(x_{k})||^2 \\ \notag
&+3\eta^2t\sum_{k=t_0-T}^{t-1+t_0} M^22L^2\eta^2T^2
||\sum_{m=1}^M\nabla f(x_{k-\tau_{k,m}})||^2\\ \notag
 &+3\eta^2t\sum_{k=t_0}^{t-1+t_0} M^22L^2\eta^2||\sum_{j=k-\tau^{max}_{k}}^{k-1}\sum_{m=1}^M\zeta_{j,m}||^2\\ \notag
  \overset{(b)}{\leq} &3\eta^2 \sum_{k=t_0}^{t-1+t_0} M^2t||\nabla f(x_{k})||^2 \\ \notag
&+3\eta^2t\sum_{k=t_0-2T}^{t-1+t_0} M^22L^2\eta^2T^3
M^2||\nabla f(x_{k})||^2\\ \notag
 &+3\eta^2t\sum_{k=t_0}^{t-1+t_0} M^22L^2\eta^2||\sum_{j=k-\tau^{max}_{k}}^{k-1}\sum_{m=1}^M\zeta_{j,m}||^2\\ \notag
\end{align}

In (a), we use the estimation for $T_1$ in the previous section, and (b) is from $\tau_{k,m}\leq T$ such that
$$\sum_{k=t_0-T}^{t-1+t_0}||\sum_{m=1}^M\nabla f(x_{k-\tau_{k,m}})||^2\leq \sum_{k=t_0-2T}^{t-1+t_0}TM^2||\nabla f(x_{k})||^2$$

\QEDA

\section*{F Proof of Lemma \ref{l3}}
Supposing there is a $\tau \leq T_{max}$, such that $||x(k+\tau)-x(k)||^2\geq S^2)$, with probability at least $1-TT_{max}e^{-T_{max}}-\frac{1}{48}$, we have

\begin{align}
\sum_{k=t_0}^{T_{max}-1+t_0}(1+2L^2\eta^2M^2T^3) ||\nabla f(x_{k})||^2\ &\geq \sum_{k=t_0}^{\tau-1+t_0}(1+2L^2\eta^2M^2T^3) ||\nabla f(x_{k})||^2 \\ \notag
&\geq \frac{||  x_{t_0+\tau-1}-x_{t_0}||^2-3\eta^2||\sum_m \sum_{i=t_0}^{t_0+\tau-1}\zeta_{i,m}||^2}{3\eta^2M^2T_{max}}\\ \notag
&-\sum_{k=t_0-2T}^{t_0-1}2L^2\eta^2T^3||\sum_{m=1}^M\nabla f(x_{k})||^2\\ \notag
&-\sum_{k=t_0}^{T_{max}-1+t_0} 2L^2\eta^2||\sum_{j=k-\tau_{k,\mu}}^{k-1}\sum_{m=1}^M\zeta_{j,m}||^2\\ \notag
&\overset{(a)}{\geq} \frac{S^2-3\eta^2M\sigma^2 T_{max}c^2c_2}{3\eta^2M^2T_{max}}-2L^2\eta^2T^3M^2 F\\ \notag
&- c2T_{max}2L^2M\eta^2T\sigma^2\\ \notag
&\geq 2T F_2\\ \notag
\end{align}
$2L^2\eta^2M^2T^3= 2L^2\eta^2M^2T^2\cdot T \leq T$. We have
\begin{equation}
\sum_{k=t_0}^{T_{max}-1+t_0} ||\nabla f(x_{k})||^2 \geq F_2
\end{equation}
In (a) we use Lemma \ref{lsum}, and with probability at least $1-TT_{max}e^{-T_{max}}$
$$\sum_{k=t_0}^{T_{max}-1+t_0} 2L^2\eta^2||\sum_{j=k-\tau^{max}_{k}}^{k-1}\sum_{m=1}^M\zeta_{j,m}||^2\leq c\sigma^22T_{max}2L^2M\eta^2T\leq c\sigma^2$$
With probability at least $1-1/48$, we have
$$||\sum_m \sum_{i=t_0}^{t_0+\tau-1}\zeta_{i,m}||^2\leq c^2M\sigma^2T_{max}c_2$$
This is due to Lemma \ref{ll1} and $2(d+1)e^{-c_2}=\frac{1}{48}$, $c_2=\log96 +\log (d+1)$ and $e^{-T_{max}+\log T+\log T_{max}}\leq 1/48$. Our claim follows.
\QEDA

\section*{G Proof of theorem \ref{st}}
In order to analyze the $x(t)$ under the updating rules, as in \cite{jin2017how},
the standard proof strategy to consider two sequences $\{x_1(t)\}$ and $\{x_2(t)\}$ as  two separate runs of algorithm \ref{a2} starting from $x(k)$(for all $t\leq k$, $x_1(t)=x_2(t)$ ).
They are coupled, such that for the Gaussian noise $\zeta_1(t)$ and $\zeta_2(t)$ in algorithm \ref{a2}, $e_1^T\zeta_1=-e_1^T\zeta_2$, where $e_1$ is the eigenvector corresponding to the minimum eigenvalue of $\nabla^2f(x)$,
and the components at any direction perpendicular to $e_1$ of $\zeta_1$ and $\zeta_2$ are equal. Given  coupling sequence  $\{x_1(k+t)\}$ and $\{x_2(k+t)\}$, let $x(t)=x_1(k+t)-x_2(k+t)$. We have
\begin{equation}
\begin{aligned}
x(k)=x(k-1)+\eta(\sum_{m=1}^M\nabla f(x_1(k-\tau_{k,m})) +\zeta_{1,k,m}-\sum_{m=1}^M\nabla f(x_2(k-\tau_{k,m})) -\zeta_{2,k,m}).
\end{aligned}
\end{equation}
We have $\nabla f(x_1)-\nabla f(x_2)=\int_0^1 \nabla^2f(tx_1+(1-t)x_2)(x_1-x_2)dt =[\nabla^2f(x_0)+\int_0^1 \nabla^2f(tx_1+(1-t)x_2)dt -\nabla^2f(x_0)](x_1-x_2) $.
Let $\boldsymbol{H}=\nabla^2f(x_0)$, $\Delta_{x_1,x_2} = \int_0^1 \nabla^2f(tx_1+(1-t)x_2)dt-\nabla^2f(x_0)$. We have

\begin{equation}\label{eq}
\begin{aligned}
x(k)=&x(k-1)+\eta[\sum_{m=1}^M (\boldsymbol{H}+\Delta_{x_1(k-\tau_{k,m}),x_2(k-\tau_{k,m})})x(k-\tau_{k,m}) +\zeta_{1,k,m}- \zeta_{2,k,m}].\\
\end{aligned}
\end{equation}

Now we want to estimate the probability of event
$$\{\max_{t\leq T_{max}}(||x_1(k+t)-x_1(k)||^2,||x_2(k+t)-x_2(k)||^2)\geq S^2\  \text{or} \ ||x(t)||\geq 2S\}$$
It is enough to consider a random variable $x'$ such that $x'(t)|E-x(t)|E=0$ where $E$ is the event $\{ \forall t \leq T_{max} :
\max_{t\leq T_{max}}(||x_1(k+t)-x_1(k)||^2,||x_2(k+t)-x_2(k)||^2)\leq S^2\}$.
This is from
\begin{equation}
\begin{aligned}
P(\forall t \leq T_{max} &\max(||x_1(k+t)-x_1(k)||^2,||x_2(k+t)-x_2(k)||^2)\leq S^2\  \text{or}\ ||x(t)||< 2S )\\
&=P(\forall t \leq T_{max} \max(||x_1(k+t)-x_1(k)||^2,||x_2(k+t)-x_2(k)||^2)\leq S^2\  \text{or}\ ||x'(t)|| < 2S ).
\end{aligned}
\end{equation}
Then we can turn to consider $x'$, such that\\

\begin{equation}
\begin{aligned}
x'(k)=x'(k-1)+\eta[\sum_{m=1}^M (\boldsymbol{H}+\Delta'_{x_1(k-\tau_{k,m}),x_2(k-\tau_{k,m})})x(k-\tau_{k,m}) +\zeta_{1,k,m}- \zeta_{2,k,m}].\\
\end{aligned}
\end{equation}

If $\max(||x_1(t)-x(k)||^2,||x_2(t)-x(k)||^2)\leq S^2$
$$\Delta'(t)=\Delta,$$
else $$\Delta'(t)=\rho S.$$
Then  $\Delta'_{x_1(k-\tau_{k,m}),x_2(k-\tau_{k,m})}\leq \rho S$. In order to simplify symbols, we denote $x=x'$.

To show that$||x(T_{max})||\geq 2S$, we consider Eq.(\ref{eq}). Let $\{\zeta_{1,i}, \zeta_{2,i}\}$, $\{\xi_{1,i,m}, \xi_{2,i,m}\}$
be the Gaussian noise and stochastic gradient noise in two runs. We set $\zeta_i=\zeta_{1,i}-\zeta_{2,i}$, $\xi_{i,m}=\xi_{1,i,m}-\xi_{2,i,m}$.
It is easy to see that $\zeta_i=2\boldsymbol{P}\zeta_{1,i}$, where $\boldsymbol{P}$ is the projection matrix to $e_1$. This is from the definition of the coupling sequence. And from Assumption \ref{a4},
$\xi_{i,m}$ is $\ell^2||x(i-\tau_{i,m})||^2$-norm-sub-Gaussian.

Then there is a polynomial function $f(t_0,t,y)$ such that $x(k)=\psi(k)+\phi(k)+\phi_{sg}(k)$
\begin{equation}
\begin{aligned}
&\psi(k)=\sqrt{M}\eta \sum_{i=0}^{k-1} f(i,k,\boldsymbol{H})\zeta_i \\
&\phi(k)=\eta \sum_m \sum_{i=0}^{k-1} f(i,k,\boldsymbol{H})\Delta(i-\tau_{i,m})x(i-\tau_{i,m})\\
&\phi_{sg}(k)=\eta \sum_m\sum_{i=0}^{k-1} f(i,k,\boldsymbol{H})\xi_{i,m}
\end{aligned}
\end{equation}
$f(t_0,t,\boldsymbol{H})$ is the solution (fundamental solution)  of the following linear equation
\begin{equation}
\begin{aligned}
x(k)&=x(k-1)+ \eta[\sum_{m=1}^m \boldsymbol{H}x(k-\tau_{k,m})]\\
x(t_0)&=\boldsymbol{I}\\
x(n)&=\boldsymbol{0} \text{ for all $n<t_0$}
\end{aligned}
\end{equation}
This is an easy inference for linear time-varying systems. And if the maximal eigenvalue of $\boldsymbol{H}$ is $\gamma$,
it is easy to see for any vector $V=\bm{P}V$ with $||V||=1$, $||f(t_0,t,\boldsymbol{H})V||_2= f(t_0,t,\gamma)\triangleq f(t_0,t)$.

\begin{lem}\label{llm}
Let $f(t_0,t)=f(t_0,t,\gamma)$ , $\beta^2(k)=\sum_{i=0}^k f^2(i,k)$ we have
\begin{enumerate}[(1)]
\item $f(t_0,t_1)f(t_1,t_2)\leq f(t_0,t_2)$
\item $f(t_1,t_2)\geq f(t_1,t_2-1)$
\item $f(k,t)\beta(k)=\sqrt{\sum_{j=0}^{k-1}f^2(k,t)f^2(j,k)}\leq \sqrt{\sum_{j=0}^{k-1}f^2(j,t)}\leq \beta(t)$
\item $f(k,t+1) \geq (1+M\eta\gamma e^{-(T+1)M \eta\gamma} )f(k,t)$ if $t-k \geq T$.
\item $q= M\eta\gamma e^{-(T+1)M \eta\gamma}$, $\beta^2(k)\geq \sum_{j=0}^{k-T} (1+q)^{2j} \geq \frac{(1+q)^{2(k-T)}}{3\cdot 2q}$ when $k-T\geq \ln2/q$
\end{enumerate}

\end{lem}
{\bf Proof:}
The first three inequalities are trivial. (4) is from Corollary \ref{f1}. (5) is easily deduced from (4).
\QEDA

Now we can estimate $\phi$ term.
\begin{equation}
\phi(t+1) = \eta \sum_m \sum_{n=0}^{t} f(n,t+1,\boldsymbol{H}) \Delta(n-\tau_{n,m})  x(n-\tau_{n,m})
\end{equation}
To give a estimation, we need the Chernoff bound. Let
$$\boldsymbol{Y}=\begin{bmatrix}

       0 & X^T \\

       X & 0

  \end{bmatrix}\text{ , }
\boldsymbol{Y}_N=\begin{bmatrix}

       0 & \psi^T \\

       \psi & 0

  \end{bmatrix}\text{ , }
\boldsymbol{Y}_\phi=\begin{bmatrix}

       0 & \phi^T \\

       \phi & 0

  \end{bmatrix}\text{ , }
  \boldsymbol{Y}_{sg}=\begin{bmatrix}

       0 & \phi_{sg}^T \\

       \phi_{sg} & 0

  \end{bmatrix}  $$

We have

\begin{theo}
For all $0\leq t\leq T_{max}$, and $\theta^2\leq \frac{1}{48\cdot c(\sum_{j=1}^{t} p^j)^2 \beta^2(t)M\eta^2 4r^2/d}$, $C_2=3\times65$, we have

\begin{equation}
\begin{aligned}
&\mathbb{E} tr\{ e^{\theta \boldsymbol{Y}_\phi(t)+\theta \boldsymbol{Y}_{sg}(t) }\}\leq  e^{c\theta^2 (\sum_{j=1}^{t} p^j)^2 \beta^2(t)M\eta^2 4r^2/d}(d+1)\\
&\mathbb{E} tr\{ e^{\theta \boldsymbol{Y}(t)}\} \leq e^{c\theta^2 (1+\sum_{j=1}^{t} p^j)^2\beta^2(t) 4M\eta^2 r^2/d}(d+1)\\
&\mathbb{E} e^{\theta^2 ||\boldsymbol{Y}(t)||^2} \leq e^{C_2c\theta^2 (1+\sum_{j=1}^{t} p^j)^2\beta^2(t) 4M\eta^2 r^2/d}(d+1)\\
\end{aligned}
\end{equation}
\end{theo}

{\bf Proof:}

We use mathematical induction.

For $t=0$, the first inequality is obviously true. For the second one
$$x(0)=\psi(0)+\phi(0)+\phi_{sg}(0)=\psi(0)$$
so since $\psi$ is sub-Gaussian, we have
\begin{equation}
\begin{aligned}
\mathbb{E} tr\{e^{\theta \boldsymbol{Y}(0)}\}&\leq e^{c\theta^2 \beta^2(0)M\eta^24r^2/d}(d+1)\\
\mathbb{E} e^{\theta^2 ||\boldsymbol{Y}(0)||^2}&\leq e^{65c\theta^2 \beta^2(0)M\eta^24r^2/d}\\
&\leq e^{C_2c\theta^2 [1+\sum_{j=1}^{0} p^j)]^2 \beta^2(0)M\eta^2 4r^2/d}
\end{aligned}
\end{equation}

Then supposing the lemma is true for all $\tau\leq t$, we consider $t+1$.
$$\mathbb{E} tr\{e^{\theta \boldsymbol{Y}_\phi(t+1)}\}=\mathbb{E} tr\{e^{\theta( \eta \sum_m \sum_i  f(i,t,H) \Delta(i-\tau_{i,m})  \boldsymbol{Y}(i-\tau_{i,m}))}\} $$

so we have
\begin{equation}
\begin{aligned}
\mathbb{E} tr\{ e^{\theta \boldsymbol{Y}_\phi(t+1) }\} &=\mathbb{E} tr \{e^{ \theta( \eta \sum_m \sum_i f(i,t+1,\boldsymbol{H}) \Delta(i-\tau_{i,m})  \boldsymbol{Y}(i-\tau_{i,m}))}\} \\
&\overset{(1)}{\leq}e^{c\theta^2 (Mt f(i,t+1) \eta\rho S)^2 (1+\sum_{j=1}^t p^j)^2\beta^2(i) M\eta^2 4r^2/d}(d+1)\\
&\leq  e^{c\theta^2 (Mt  \eta\rho S)^2(1+\sum_{j=1}^t p^j)^2 \beta^2(t+1)M\eta^2 4r^2/d}(d+1)\\
\end{aligned}
\end{equation}
(1) is from lemma \ref{ll8}, $e^X=1+X+\frac{X^2}{2}+...$ $ ||f(i,t+1,\boldsymbol{H})||\leq  f(i,t+1)$, $\beta(i-\tau)\leq \beta(i)$, $ \Delta(i-\tau_{i,m})\leq \rho S$.

And we have
\begin{equation}
\begin{aligned}
\mathbb{E} tr\{ e^{\theta \boldsymbol{Y}_{sg}(t+1) }\} &=tr\{\mathbb{E} e^{ \sum_i\sum_m c\theta^2  (  f(i,t+1)\eta \ell||\boldsymbol{Y}(i-\tau_{i,m})||)^2}\boldsymbol{I}\}\\
&\leq e^{C_2c\theta^2 (\sqrt{Mt} f(i,t+1) \eta \ell)^2 (1+\sum_{j=1}^t p^j)^2\beta^2(i) M\eta^2 4r^2/d}(d+1)\\
&\leq  e^{C_2c\theta^2 (\sqrt{Mt}  \eta\ell)^2(1+\sum_{j=1}^t p^j)^2 \beta^2(t+1)M\eta^2 4r^2/d}(d+1)\\
\end{aligned}
\end{equation}
Thus
\begin{equation}
\begin{aligned}
\mathbb{E} tr\{ e^{\theta \boldsymbol{Y}_\phi(t+1)+\theta \boldsymbol{Y}_{sg}(t+1) }\}
&\leq  e^{c\theta^2 (Mt  \eta\rho S+\sqrt{C_2}\sqrt{Mt}  \eta\ell)^2(1+\sum_{j=1}^t p^j)^2 \beta^2(t+1)M\eta^2 4r^2/d}(d+1)\\
&\leq  e^{c\theta^2 p^2(1+\sum_{j=1}^t p^j)^2 \beta^2(t+1)M\eta^2 4r^2/d}(d+1)\\
&=  e^{c\theta^2 (\sum_{j=1}^{t+1} p^j)^2 \beta^2(t+1)M\eta^2 4r^2/d}(d+1)\\
\end{aligned}
\end{equation}

\begin{equation}
\begin{aligned}
\mathbb{E} tr\{ e^{\theta \boldsymbol{Y}(t+1)}\} &=\mathbb{E} tr \{e^{\theta(\boldsymbol{Y}_N(t+1)+\boldsymbol{Y}_{sg}(t+1) +\boldsymbol{Y}_\phi(t+1))}\}\\
&\leq e^{c\theta^2 (1+\sum_{j=1}^{t+1} p^j)^2\beta^2(t+1) 4M\eta^2 r^2/d}(d+1)\\
\end{aligned}
\end{equation}
As for $\mathbb{E} e^{\theta^2 ||\boldsymbol{Y}(t+1)||^2}$, using Lemma \ref{ex}, we have

\begin{equation}
\begin{aligned}
\mathbb{E} e^{\theta^2 ||\boldsymbol{Y}_N(t)||^2} \leq e^{65c\theta^2 \beta^2(t)M^2\eta^24r^2/d}\\
\end{aligned}
\end{equation}

And
\begin{equation}
\begin{aligned}
\mathbb{E} e^{\theta^2 ||\boldsymbol{Y}_{sg}(t+1)||^2} &\leq \mathbb{E} e^{\sum_i \sum_m  65c\theta^2 (f(i,t+1) \eta \ell||\boldsymbol{Y}(i-\tau_{i,m})||)^2}\\
&\leq e^{65C_2c\theta^2 (\sqrt{Mt} f(i,t+1) \eta \ell)^2 (1+\sum_{j=1}^t p^j)^2\beta^2(i) M\eta^2 4r^2/d}\\
&\leq  e^{65C_2c\theta^2 (\sqrt{Mt} \eta\ell)^2(1+\sum_{j=1}^t p^j)^2 \beta^2(t+1)M\eta^2 4r^2/d}\\
\end{aligned}
\end{equation}

\begin{equation}
\begin{aligned}
\mathbb{E} e^{\theta^2 ||\boldsymbol{Y}_\phi(t)||^2 } &=\mathbb{E} e^{ \theta^2( \eta \sum_m \sum_i f(i,t,\boldsymbol{H}) \Delta(i-\tau_{i,m})||\boldsymbol{Y}(i-\tau_{i,m})||)^2} \\
&\leq e^{65C_2c\theta^2 (Mt f(i,t+1) \eta\rho S)^2 (1+\sum_{j=1}^t p^j)^2\beta^2(i) M\eta^2 4r^2/d}\\
&\leq  e^{65 C_2c\theta^2 (Mt  \eta\rho S)^2(1+\sum_{j=1}^t p^j)^2 \beta^2(t+1)M\eta^2 4r^2/d}
\end{aligned}
\end{equation}

Thus, we have

\begin{equation}
\begin{aligned}
\mathbb{E} e^{\theta^2 ||\boldsymbol{Y}(t+1)||^2}&\leq \mathbb{E} e^{3\theta^2 ||\boldsymbol{Y}_N(t+1)||^2+3\theta^2 ||\boldsymbol{Y}_{sg}(t+1)||^2+3\theta^2 ||\boldsymbol{Y}_\phi(t+1)||^2}\\
&\leq e^{65c\theta^2 3[1+(\sqrt{C_2}Mt  \eta\rho S+\sqrt{C_2} \sqrt{Mt} \eta\ell)(1+\sum_{j=1}^t p^j)]^2 \beta^2(t+1)M\eta^2 4r^2/d}\\
&\leq e^{C_2c\theta^2 [1+\sum_{j=1}^{t+1} p^j)]^2 \beta^2(t+1)M\eta^2 4r^2/d}\\
\end{aligned}
\end{equation}
where we use (b) of Lemma \ref{ie}.
\QEDA

By Lemma \ref{pr}, we have
\begin{cor}
For any $\iota>0$
$$P(||\phi_{sg}(k)+\phi(k)||\leq \frac{\beta (k)\sqrt{M}\eta 2r}{C\sqrt{d}}(\sqrt{48c}+\iota) )\geq 1-2(d+1)e^{-\iota}$$
\end{cor}
where $\frac{1}{C}=\sum_{i=1}^{\infty}p^i=\frac{p}{1-p}$.\\

We select $\iota=b=\log(2(d+1)) +\log2$, then $\frac{2\sqrt{48c}+2b }{C}=\frac{1}{2}$, $P(||\phi_{sg}(k)+\phi(k)||\leq \frac{\beta(k)\sqrt{M}\eta r}{2\sqrt{d}})\geq \frac{1}{2}$

\begin{lem}\label{l4}
For all $k$:
$$P(||\psi(k)||\geq \frac{\beta(k)\sqrt{M}2\eta r}{3\sqrt{d}})\geq \frac{2}{3}$$
\end{lem}
{\bf Proof:}
Since $\psi$ is Gaussian, $P(|X|\leq \lambda \sigma)\leq 2\lambda/\sqrt{2\pi}\leq \lambda$ for all normal random variable X.
Let $\lambda=\frac{1}{3}$. $\psi(k)\geq \frac{\beta(k)\sqrt{M}\eta 2r}{3\sqrt{d}}$ with probability $2/3$.
\QEDA

Using these lemmas, with probability $1/6$, $||x(k)||\geq ||\psi(k)||-||\phi_{sg}(k)+\phi(k)||\geq \frac{\beta(k)\sqrt{M}\eta r}{6\sqrt{d}}$. Then we have, with probability at least $1/6$, $\max_{0\leq t \leq T_{max}} (||x_1(k+t)-x_1(k)||,||x_2(k+t)-x_2(k)||)\geq S$ or $||x_1(k+t)-x_2(k+t)||\geq \frac{\beta(k)\sqrt{M}\eta r}{6\sqrt{d}}$, and
for all $k-T\geq \ln2/q$, $q=M\eta\sqrt{\rho\epsilon}/2e^{-f}$, $f=(T+1)M\eta\sqrt{\rho\epsilon}/2$, we have $\beta^2(k)\geq \frac{(1+q)^{k-T}}{6q}$

For coupling two runs of algorithm \ref{a2}$\{x_1(t)\}$, $\{x_2(t)\}$,
\begin{equation}
\begin{aligned}
\max(||x_1(k+t)-x_1(k)||,||x_2(k+t)-x_2(k)||)\geq \frac{1}{2}||x_1(k+t)-x_2(k+t)||
\end{aligned}
\end{equation}
Since $u\leq \tilde{O}(1)$ is large enough to satisfy $(e)$ in Lemma \ref{ie} and $\eta$ small enough such that $(1+q)^{1/q}>2$, we have
$$\frac{\beta(T_{max})\sqrt{M}\eta r}{6\sqrt{d}} \geq \frac{(1+q)^{T_{max}-T}\sqrt{M}\eta r }{6\sqrt{3}\sqrt{2q d}}\geq\frac{(1+q)^{u/q }\sqrt{M}\eta r}{6\sqrt{3}\sqrt{2q d}}\geq \frac{2^{u }\sqrt{M}\eta r}{6\sqrt{3}\sqrt{2q d}}\geq 2S$$

Thus with probability at least $1/6$ $$\max_{t\leq T_{max}}(||x_1(k+t)-x_1(k)||,||x_2(k+t)-x_2(k)||)\geq S$$
so that we have
\begin{equation}
\begin{aligned}
P(\max_{t\leq T_{max}}||x_1(k+t)-x_1(k)||\geq S)&=P(\max_{t\leq T_{max}}||x_2(k+t)-x_2(k)||\geq S)\\
&\geq \frac{1}{2}P(\max_{t\leq T_{max}}(||x_1(k+t)-x_1(k)||,||x_2(k+t)-x_2(k)||)\geq S)\\
&\geq 1/12
\end{aligned}
\end{equation}
This prove Theorem \ref{st}.

\section*{H  The Growth Rate of Polynomial $f(t_1,t_2)$}\label{s11}
In this section, we will prove Theorem \ref{th8} and the last property of polynomial $f(t_1,t_2)$ in lemma \ref{llm}. Firstly, in the synchronous case, the delay $T=0$.
We know Lyapunov's First Theorem.
\begin{lem}
Let $\boldsymbol{A}$ to be a symmetric matrix, with maximum eigenvalue $\gamma>0$. Suppose the updating rules of x is
\begin{equation}\label{ll}
\begin{aligned}
x(n+1)=x(n)+\boldsymbol{A}x(n)
\end{aligned}
\end{equation}
Then x(n) is exponential unstable in the neighborhood of zero.
\end{lem}
This can be proved by choosing a Lyapunov function. We consider $V(n)=x(n)^T\boldsymbol{P}x(n)$,
where $\boldsymbol{P}$ is the Projection matrix to the subspace of the maximum eigenvalue. We can show that $V(n+1)=(1+\gamma)^2V(n)$.

This method can be generalized to the asynchronous(time-delayed) systems.
There are many works on the stability of the time-delay system by considering Lyapunov functional \cite{Kharitonov2003Lyapunov,Gu1999Discretized,Han2005On}.
Constructing a Lyapunov functional is generally tricky. One way to avoid this is to use  Razumikhin-type theorems \cite{XU1994An,Zhou2018Improved}.  A stochastic version of Razumikhin theorems is proved in \cite{Mao1999Razumikhin}. There are few works on the instability of the time-delayed system.
\cite{Haddock1996Instability} used Razumikhin-type theorems to study the instability,
and the work in \cite{Raffoul2013INEQUALITIES} constructed a Lyapunov functional. It was shown that when the delay is small enough, the system is exponentially unstable.

\subsubsection*{H.1 A Rough Estimation}
Here we give a much easier analysis for the linear time-delay system without using Lyapunov functional.
\begin{lem} \label{lm2}
Let $\boldsymbol{A}$ to be a symmetric matrix
\begin{equation}\label{lin}
\begin{aligned}
x(n+1)=x(n)+\sum_i^m \boldsymbol{A}x(n-\tau_{n,i})\\
\boldsymbol{P}x(0)\neq \boldsymbol{0}, x(t)=\boldsymbol{0} \text{ for all $t< 0$}
\end{aligned}
\end{equation}
with $0\leq \tau\leq T$, the largest eigenvalue of $\boldsymbol{A}$ is $\gamma$, $\boldsymbol{P}$ be the projection matrix to the eigenvalues $\gamma$. Let $V(n)=x(n)^T\boldsymbol{P}x(n)$, If $m\gamma-m^3\gamma^3 T^2=q>0$, we have $V(n+1)\geq (1+q) V(n)$ for $n\geq T$ and $V(n+1)\geq V(n)$ for $n< T$.
\end{lem}

Let $\boldsymbol{P}$ be the projection matrix of $\boldsymbol{A}$ to the subspace of maximum eigenvalue and $V(n)=x(n)^T\boldsymbol{P}x(n)$. We have

\begin{equation}
\begin{aligned}
V(n+1)=&x(n)^T\boldsymbol{P}x(n)+2x(n)^T\boldsymbol{P}[\sum_i^m \boldsymbol{A}x(n-\tau_{n,i})] \\
+&[\sum_i^m \boldsymbol{A} x(n-\tau_{n,i}) ]^T\boldsymbol{P}[\sum_i^m \boldsymbol{A} x(n-\tau_{n,i})]\\
\geq& V(n) + 2x(n)^T\boldsymbol{P}\sum_i^m \boldsymbol{A} x(n-\tau_{n,i})\\
\end{aligned}
\end{equation}
Let $i$ in the set $\{1,2,3...m'(n)\}$ such that $x(n-\tau_{n,i})\neq 0$ and if $n\geq T$, $m'(n)=m$. For simplicity, we use $m$ to represent $m'(n)$.
Using the fact $\left<a,b\right>=\frac{1}{2}(||a||^2+||b||^2-||a-b||^2)$, we have

\begin{align}
V(n+1)=& V(n)+ m x(n)^T\boldsymbol{P} \boldsymbol{A}x(n)+  m \frac{1}{m}\sum_i^{m} x(n-\tau_{n,i})^T\boldsymbol{P} \boldsymbol{A} \frac{1}{m}\sum_i^{m} x(n-\tau_{n,i})\\  \notag
-&m [x(n)-\frac{1}{m}\sum_i^{m} x(n-\tau_{n,i})]^T\boldsymbol{P} \boldsymbol{A} [x(n)-\frac{1}{m}\sum_i^{m} x(n-\tau_{n,i})]\\ \notag
\geq& V(n) +m \gamma V(n) +  m\frac{1}{m}\sum_i^m x(n-\tau_{n,i})^T\boldsymbol{P} \boldsymbol{A}\frac{1}{m}\sum_i^m x(n-\tau_{n,i})\\ \notag
-&m [x(n)-\frac{1}{m}\sum_i^m x(n-\tau_{n,i})]^T\boldsymbol{P}\boldsymbol{ A} [x(n)-\frac{1}{m}\sum_i^m x(n-\tau_{n,i})]\\ \notag
\geq& V(n)+ m \gamma V(n)+  m\frac{1}{m}\sum_i^m x(n-\tau_{n,i})^T\boldsymbol{P} \boldsymbol{A}\frac{1}{m}\sum_i^m x(n-\tau_{n,i})\\ \notag
-& m [\sum_{t=n-\tau_{n,\mu}}^{n-1} \sum_i^m \boldsymbol{A}  x(t-\tau_{t,i})]^T \boldsymbol{P} \boldsymbol{A} [\sum_{t=n-\tau_{n,\mu}}^{n-1} \sum_i^m \boldsymbol{A} x(t-\tau_{t,i})]\\ \notag
\geq& V(n) +m \gamma V(n) + m\gamma \frac{1}{m}\sum_i^m x(n-\tau_{n,i})^T\boldsymbol{P}\frac{1}{m}\sum_i^m x(n-\tau_{n,i})\\ \notag
-& m\gamma^3 T\sum_{t=n-T}^{n-1} ( \sum_i^m  x(t-\tau_{t,i}))^T\boldsymbol{P}( \sum_i^m  x(t-\tau_{t,i}))\\ \notag
\geq& V(n) + m  \gamma V(n)\\ \notag
+& m\gamma \frac{1}{m}\sum_i^m x(n-\tau_{n,i})^T\boldsymbol{P}\frac{1}{m}\sum_i^m x(n-\tau_{n,i})\}\\ \notag
-& m^2\gamma^3 T\sum_{t=n-T}^{n-1}\sum_i^m   V(t-\tau_{t,i})
\end{align}

Note that from Eq.(\ref{lin}), since $\gamma>0$ $||\boldsymbol{P}x(n)||$ will keep increasing.
$V(n)\geq V(n-\tau)$ for all $\tau\geq 0$. If $m\gamma-m^3\gamma^3 T^2=q>0$, we have $V(n+1)\geq (1+q)V(n)$, $||\boldsymbol{P}x(n+1)||\geq \sqrt{1+q} ||\boldsymbol{P}x(n)||$ if $n>T$.

\subsubsection*{H.2 Razumikhin-Lyapunov Method}
$q=m\gamma-m^3\gamma^3 T^2$, even when $T=0$, $q\leq m\gamma$. But we know that  $V(n+1)=(1+2m\gamma +m^2\gamma^2)V(n)$,
so that $1+q$ is a very rough estimation. Here, using Razumikhin technique,
we give a new theorem to get a better estimation and it can go beyond $T\sim \frac{1}{\gamma}$ cases ($m\gamma-m^3\gamma^3 T^2>0$),
This theorem is inspired by the proof in \cite{mao1996razumikhin,Mao1999Razumikhin}.

\begin{theo} \label{tt4}(Restatement of Theorem \ref{th8}) For a discrete system,
$V(n,x)$ is a positive value Lyapunov function. Let $\Omega$ be the space of discrete function $x(\cdot)$ from $\{-T,...0,1,2,...\}$
to $\mathbb{R}$ and $x(\cdot)$ is a solution of the given discrete system equation. Suppose there exit $q, q_m$ satisfying the following two conditions
\begin{equation}\label{con1}
\begin{aligned}
&(a)V(t+1,x(t+1))\geq q_m V(t,x(t)), q_m>0 \text{ (Bounded difference condition.) } \\
&(b)\text{If }\ V(t-\tau,x(t-\tau))\geq (1+q)^{-T} \frac{q_m}{1+q}V(t,x(t)) \forall 0\leq \tau \leq T \\
&\text{then }\ V(t+1,x(t+1))\geq (1+q) V(t,x(t))\text{ (Razumikhin condition.) }
\end{aligned}
\end{equation}
Then for any $x(\cdot)\in \Omega$ satisfying  that for all $-T\leq t\leq 0 $, $V(t,x(t))\geq p V(0,x(0))$ with $0<p\leq 1$, we have $V(t,x(t))\geq (1+q)^t pV(0,x(0))$ for all $t>0$.
\end{theo}

{\bf Proof:}

Let $B(n)=(1+q)^{-n} V(n)$. In order to prove our theorem, we only need to show $B(n)$ have a lower bound.

$B(0)=V(0)\geq p V(0)\triangleq p'$. Assuming there is a $t>0$ such that $B(t)=(1+q)^{-t} V(t)< p' $, select the minimum one as $t$,
such that $B(k)\geq p'$ for all $k<t$, and $B(t)< p'$. Note that $V(t)\geq q_m V(t-1)$ so that $B(t)\geq p'\frac{q_m}{1+q}$. Then for all k satisfying $t-T\leq k\leq t$,
\begin{equation}
\begin{aligned}
V(k)&=(1+q)^{k}B(k)\geq (1+q)^k p'\frac{q_m}{1+q}= (1+q)^{k-t} (1+q)^{t} p'\frac{q_m}{1+q}, \\
&\geq (1+q)^{k-t} (1+q)^{t} \frac{q_m}{1+q} B(t)\geq (1+q)^{-T}\frac{q_m}{1+q} V(t).
\end{aligned}
\end{equation}
So that we have $V(t+1)\geq (1+q) V(t) $, $ B(t+1)\geq B(t)\geq p' \frac{q_m}{1+q} $. If $B(t+1)\geq p'$, $V(t+2)\geq q_m V(n+1)$,
so that $B(t+2)\geq B(t+1) \frac{q_m}{1+q}\geq p'\frac{q_m}{1+q}$. If $B(t+1)<p'$, $V(t+1-\tau)\geq (1+q)^{-T}\frac{q_m}{1+q}V(t+1)$,
from the condition in (\ref{con1}), $ B(t+2)\geq B(t+1)\geq p' \frac{q_m}{1+q} $. This process can be continued, such that $B(t) \geq p'\frac{q_m}{1+q} $ for any $t$. Our claim follows.
\QEDA

Using Theorem \ref{tt4} to (\ref{lin}), we set $ V(n,x)=||\boldsymbol{P}x(n)||$.  Supposing $x(0)=I,x(-t)=0$ for all $t>0$, $||\boldsymbol{P}x(t)||=e_1^Tx(t)$ and $q_m=1$. As shown in  section \ref{eia}, we have:
\begin{cor}\label{ff}
Let $f(k,t)$ be the polynomial in lemma \ref{llm}. $f(k,t+1)\geq (1+q) f(k,t)$ if $t-k\geq T$, where $q=M\eta \gamma e^{-(T+1)M\eta\gamma}$.
\end{cor}

\end{document}